\documentclass[11pt]{article}

\PassOptionsToPackage{table}{xcolor}
\usepackage[preprint]{acl}

\usepackage{times}
\usepackage{latexsym}
\usepackage[T1]{fontenc}
\usepackage[utf8]{inputenc}
\usepackage{microtype}
\IfFileExists{inconsolata.sty}{\usepackage{inconsolata}}{}
\usepackage{graphicx}

\usepackage{amsmath}
\usepackage{amssymb}
\usepackage{booktabs}
\usepackage{multirow}
\usepackage{array}
\usepackage{xcolor}
\usepackage{tikz}
\usetikzlibrary{arrows.meta,positioning,fit,calc,backgrounds,patterns}
\usepackage{algorithm}
\usepackage{algorithmic}
\usepackage{xspace}
\usepackage{enumitem}

\newcommand{\method}{FlexA-Mem\xspace}
\newcommand{\sumret}{Summary+A-Mem\xspace}

\definecolor{firstbg}{RGB}{200,222,250}
\definecolor{secondbg}{RGB}{253,224,190}
\definecolor{thirdbg}{RGB}{214,238,206}
\newcommand{\first}[1]{\cellcolor{firstbg}#1}
\newcommand{\second}[1]{\cellcolor{secondbg}#1}
\newcommand{\third}[1]{\cellcolor{thirdbg}#1}
\newcommand{\unused}[1]{\textcolor{gray}{#1}}
\newcommand{\rec}{\mathrm{rec}}
\newcommand{\exc}{\mathrm{exc}}
\newcommand{\raw}{\mathrm{raw}}
\newcommand{\len}[1]{\lvert #1 \rvert}

\title{FlexA-Mem: Budget-Adaptive Agentic Memory for LLM Agents}
\author{
  \textbf{Wujiang Xu}\textsuperscript{1} \quad
  \textbf{Jiaojiao Han}\textsuperscript{2} \quad
  \textbf{Xi Zhu}\textsuperscript{1} \quad
  \textbf{Qizheng Zhang}\textsuperscript{3} \quad 
  \textbf{Kaiqu Liang}\textsuperscript{4} \quad \\
  \textbf{Yutong Yin}\textsuperscript{5} \quad
  \textbf{Kai Mei}\textsuperscript{1} \quad
  \textbf{Dimitris N. Metaxas}\textsuperscript{1} \\
  \textsuperscript{1}Rutgers University \textsuperscript{2}Independent Researcher \textsuperscript{3}Stanford University \\
  \textsuperscript{4}Pricenton University
  \textsuperscript{5}Northwestern University
}

\begin{document}
\maketitle

\begin{abstract}
Large language model (LLM) agents work in scenarios with very different memory needs. A question about past
conversations needs the exact facts it names, while a multi-step task needs a trace of the steps already taken
together with details that the next step cannot name. Existing memories are each designed and tested for one kind
of scenario. Retrieval memories such as A-Mem miss steps that the current query does not express, summaries drop
exact strings, and pairing the two fixes the split of the budget in advance. We propose \method, a general memory
that writes each unit of history once into an immutable store and renders a fresh memory for every call. The
rendered memory is a compact timeline of records plus detail that a fixed rule fits into the remaining room without
an LLM call. The detail is either relevant units as raw text or excerpts or the verbatim pieces that best match the
query. With one store, \method serves both kinds of scenario. On AppWorld challenge tasks at 4k, it outperforms the second-best method by 11.5\%, and on 147 unused
tasks with predictions fixed in advance it outperforms A-Mem with a summary by 12.2\%. On LongMemEval and LoCoMo, it matches a dedicated keyword retriever that writes no
memory, while memories built on text written by an LLM fall far behind at tight budgets. Ablations show that the
timeline carries AppWorld completion, whereas recovery mainly saves steps there and decides accuracy on exact facts.
Code is available at \url{https://github.com/WujiangXu/Flex-Amem}.
\end{abstract}

\section{Introduction}
\label{sec:intro}

Large language model (LLM) agents now work in many settings. They call the APIs of apps \citep{trivedi2024appworld},
resolve software issues by editing and testing code \citep{jimenez2024swebench,yang2024sweagent}, and answer
questions about conversations held over dozens of sessions \citep{wu2025longmemeval}. In each setting the history
soon outgrows a single call, and models use the middle of a long context less reliably \citep{liu2024lost}. Therefore, each call works under a fixed token budget, and memory decides what fills it.

What should fill the budget depends on the scenario. A question about a past conversation usually
names its evidence and depends on a few exact facts, such as a date mentioned weeks before. A multi-step task depends
first on an overview of what the agent has done. An AppWorld agent that sends money must log in, list the APIs and
read their documentation, and if these steps leave its context it starts them again. It also needs exact details that change from step to step, such as an API argument or an earlier output. Some calls need all of a few past units, and others need a little of every past unit.

Existing memories are each designed for one of these needs and tested on benchmarks of the matching kind. Retrieval
memories such as A-Mem \citep{xu2025amem} return the notes closest to the query, which suits fact recall, but an
agent asked to send money has no reason to search for its earlier login. Compaction replaces old observations with
placeholders or an LLM summary \citep{wang2025openhands,lindenbauer2025complexity}, which keeps an overview of a long
trajectory but drops the exact strings a fact question asks for. \sumret combines the two with a fixed split of the
budget, and MemGPT and Sculptor leave hidden details to tools \citep{packer2023memgpt,li2026sculptor} that agents rarely
call \citep{fan2026harness,li2026acm}. As a result, each method works well in its own scenario but loses much of its strength in the other.

\begin{figure}[t]
  \centering
  \IfFileExists{figures/intro_memory.pdf}{%
    \includegraphics[width=\columnwidth]{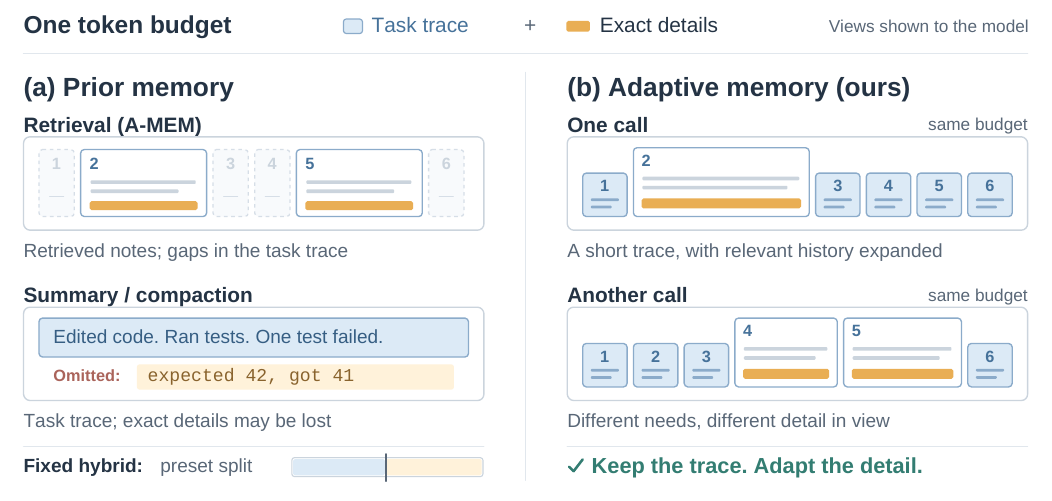}%
  }{%
    \fbox{\parbox[c][3.6cm][c]{0.94\columnwidth}{\centering\small\color{gray}
      Introduction figure (drawn by the authors).}}%
  }
  \vspace{-20pt}
  \caption{Retrieval suits fact recall but loses the trace of a multi-step task, a summary keeps the trace but loses exact facts, and \method serves both needs within a single token budget.}
  \label{fig:intro}
  \vspace{-5pt}
\end{figure}

A memory that generalizes across scenarios is hard to build because the right balance between overview and detail
is not known when the history is written (Figure~\ref{fig:intro}). It depends on the scenario, on the step within a
task and on the budget of the call. Nor can the query alone settle it. With its task and latest steps as the query,
keyword matching finds the details an AppWorld agent needs no better than chance. For this reason, a memory that commits at write time, by summarizing, by evolving notes or by fixing a split, is
tuned to one scenario by construction. The
decision has to be made at every call, and it must stay cheap, because an agent calls the model at every step.

To this end, we propose \method, a general memory that writes each unit of history once and renders it anew at every
call. The store keeps the raw text of every unit, and each rendering shows a timeline of short records that does not
depend on the query. It fills the remaining room with recovered detail by a fixed rule, without an LLM call. Our
contributions are summarized as follows.
\begin{itemize}[leftmargin=*,itemsep=1pt,topsep=2pt,parsep=0pt]
\item We introduce a write-once, render-per-call design for agent memory. It stores every unit of history immutably
and decides only at read time how much of each unit a call sees. Nothing is discarded before the scenario, the
query and the budget are known.
\item We propose fit-or-degrade, a budget-aware renderer that needs no LLM call. It gives each relevant unit the
highest resolution that fits, raw text, an excerpt or its record, and turns more units into excerpts as the budget
shrinks. As a drop-in renderer it raises the AppWorld completion of A-Mem's own notes by 10.7\%.
\item We show that one store serves both kinds of scenario. \method outperforms the second-best method by 11.5\% on AppWorld challenge tasks and the strongest
baseline by 12.2\% on 147 unused tasks, and on fact recall it matches a dedicated keyword retriever. In our ablations the timeline carries agent completion, while recovery saves steps there and decides
accuracy when the answer is an exact fact.
\end{itemize}

\section{Related Work}
\label{sec:related}

Memory is essential for personalized agents~\citep{liang2026learning,iagent}, enabling them to tailor their services to individual users' needs and preferences.
Most agent memories store past interactions and retrieve a few at each call
\citep{park2023generative,zhong2024memorybank,chhikara2025mem0,rasmussen2025zep}. A-Mem links its notes with an LLM
and lets new notes rewrite their neighbors \citep{xu2025amem}, and unretrieved steps are absent from the call. A
second line keeps the history at several levels of detail
\citep{packer2023memgpt,sarthi2024raptor,rezazadeh2025memtree,xu2026memgas}. MemGPT, RAPTOR and MemTree fix the level at write time, while MemGAS and SCM choose it with an LLM
\citep{wang2025scm} and Nemori adds raw messages of top episodes \citep{ma2026nemori}. The closest designs choose a
form per item under a budget \citep{liao2026ace,kang2026retainconsolidate}. ReadAgent lets
an LLM swap gists for raw pages \citep{lee2024readagent}, and AFM falls back from full to compressed to placeholder
forms while packing oldest first \citep{cruz2025afm}.  A third line lets the agent manage its context
through tools \citep{hu2025hiagent,li2026sculptor,ehrlich2026lcm,li2026acm} or learns this with reinforcement learning \citep{sun2025contextfolding,ye2026agentfold,zhou2026mem1,yan2026memoryr1},
but agents seldom act on their own \citep{fan2026harness,li2026acm}.

\method builds on the two closest designs, taking degradation within a budget from AFM and in-place replacement of a
gist by raw text from ReadAgent. Keeping raw text, several resolutions and in-place replacement are each known. The new design makes these choices at read time by a fixed rule, without an LLM call or an action of the agent.
Unlike A-Mem it never rewrites a note, and unlike AFM it keeps a record of every unit and visits units by relevance
rather than age. Our experiments then show that one
store serves both kinds of scenario. Appendix~\ref{app:related} covers context compression and other memory systems in more detail.

\section{FlexA-Mem}
\label{sec:method}

\begin{figure*}[t]
  \centering
  \IfFileExists{figures/framework.pdf}{%
    \includegraphics[width=\textwidth]{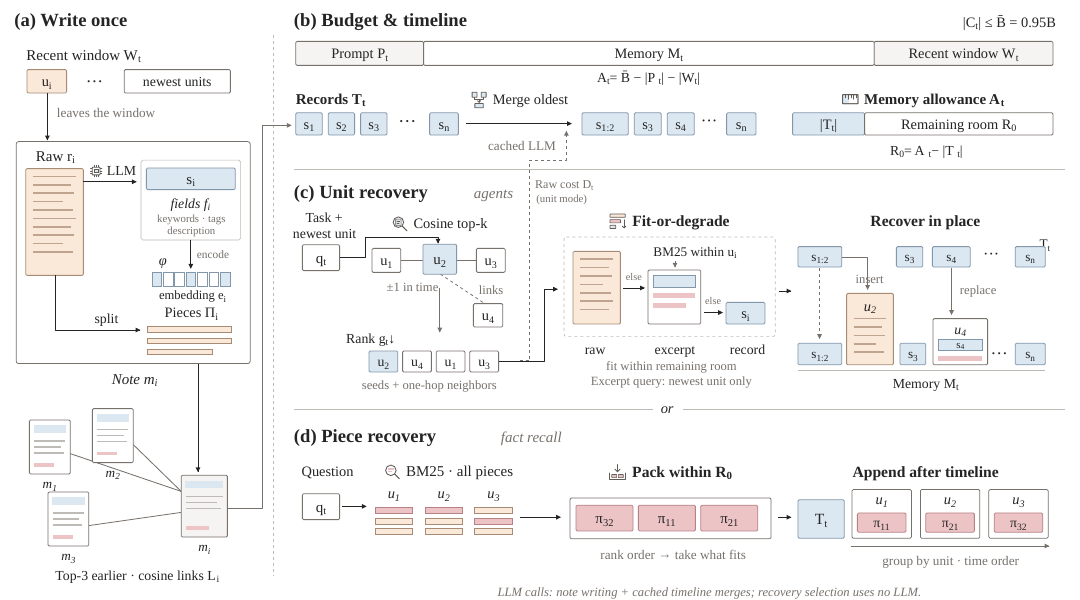}%
  }{%
    \fbox{\parbox[c][4.4cm][c]{0.96\textwidth}{\centering\small\color{gray}
      Framework figure (drawn by the authors).}}%
  }
  \vspace{-20pt}
  \caption{Overview of \method. Each unit is written once and keeps its raw text in the store. Every call shows a
  compact timeline of records and fills the rest of the budget by unit or piece recovery.}
  \vspace{-5pt}
  \label{fig:framework}
\end{figure*}

\method writes each unit once into an immutable store and renders a fresh memory for every call from one timeline
and one of two recovery modes
(Figure~\ref{fig:framework}). Call $t$ of an agent or a reader sees a context
\begin{equation}
  C_t = [\,P_t;\ M_t;\ W_t\,], \qquad |C_t| \le \bar{B},
  \label{eq:context}
\end{equation}
where $P_t$ is the fixed part of the prompt (instructions and the task or question) and $M_t$ is the memory. The
recent window $W_t$ holds the newest units shown unchanged. Lengths are counted in tokens with one shared tokenizer, and
the prompt length $\bar{B} = 0.95B$ for a budget $B$ leaves a small safety margin. The allowance
$A_t = \bar{B} - |P_t| - |W_t|$ is the room left for memory. Memory is kept in units. For an agent, consecutive steps,
each an action with the environment's output, are packed into one unit until it reaches 1{,}000 tokens. For a reader
that answers a question about a past conversation, each session is cut at user turns into spans of at most about
2{,}000 tokens, and there is no recent window. The units older than the recent window are $u_1, \dots, u_n$ in time
order.

\subsection{Writing Memory}
\label{sec:write}

When a unit $u_i$ leaves the recent window, \method writes it once as a note
$m_i = (r_i, s_i, f_i, e_i, L_i, \Pi_i)$. The raw text $r_i$ is the unit exactly as it occurred and never changes. One LLM call reads $r_i$ and writes the record
$s_i$, at most 80 words, with retrieval fields $f_i$ made of keywords, tags and a short description. A small sentence
encoder $\phi$ gives the embedding $e_i = \phi(s_i \Vert f_i)$, where $\Vert$ denotes concatenation, and the links $L_i$
connect $u_i$ in both directions to the three earlier notes with the highest cosine similarity to $e_i$. Without an
LLM call, \method also cuts $r_i$ at line breaks into pieces $\Pi_i = (\pi_{i1}, \dots, \pi_{iK_i})$ of at most
$\kappa$ tokens. Each piece stays inside one conversation turn or one block of an agent step. \method never rewrites a note, since a later call recovers detail from the exact raw text and a rewritten neighbor would add cost but no detail.

\subsection{The Timeline of Records}
\label{sec:timeline}

Every call shows the timeline $T_t$, the records of past units in time order, which keeps a compact trace of what the
agent has done. When it grows too long, \method replaces up to four consecutive records, oldest first, with a merged
record that one LLM call writes. Let $\|T_t\|$ be the length of the timeline, $\rho$ the share of the allowance given
to it, and $D_t$ the total cost of showing every candidate of unit recovery as raw text (Section~\ref{sec:recovery}).
Merging repeats, oldest first, while
$\|T_t\| > \max(\rho A_t, A_t - D_t)$. A merged record is computed once and reused by later calls. Merging loses the detail of separate
records in the context, but the raw text of every unit stays in the store and can still be recovered. On agentic tasks
the oldest records are also hidden for one call when the newest unit alone takes more than a quarter of the budget
(Appendix~\ref{app:impl}). The room left for recovery is $R_0 = A_t - \|T_t\|$.

\subsection{Unit Recovery}
\label{sec:recovery}

Unit recovery shows each unit at one of three resolutions, its record, an excerpt or its raw text, and a recovered
unit appears in its place in the timeline. Candidates come from the query embedding $\phi(q_t)$. For a reader it embeds the question,
and for an agent it is the normalized sum of the embeddings of the task and of the newest unit
(Table~\ref{tab:queries} in Appendix~\ref{app:impl}). Let $c_t(j)$ be the cosine similarity between $\phi(q_t)$ and $e_j$, let the seeds $S_t$ be
the $k$ units with the highest $c_t$, and let $N_i$ hold the two units adjacent to $u_i$ in time. A unit $u_j$ receives
the score
\begin{equation}
  \begin{aligned}
  g_t(j) = \max \bigl\{ &\lambda^{[i \neq j]}\, c_t(i) \;:\\
  &i \in S_t,\ j \in \{i\} \cup L_i \cup N_i \bigr\},
  \end{aligned}
  \label{eq:score}
\end{equation}
where $[i \neq j]$ is 1 when $u_j$ is reached from another seed and 0 otherwise, and $\lambda < 1$ discounts units
reached through a link or through time. The candidates are the seeds and the units they reach, visited in decreasing
order of $g_t$. Links recover a unit that is connected to what the query concerns, even if its own text does not resemble the query.

Let $x_i^{v}$ denote the text of $u_i$ at resolution $v$, and let $\tau_i = 1$ if the record of $u_i$ still stands
alone in the timeline and $\tau_i = 0$ if it has been merged. Showing $u_i$ at resolution $v$ costs
\begin{equation}
  \mathrm{cost}_i(v) =
  \begin{cases}
    0 & \text{if } v = \rec,\\
    \len{x_i^{v}} - \tau_i \len{s_i} & \text{otherwise}.
  \end{cases}
  \label{eq:cost}
\end{equation}
A recovered unit takes the place of its own record, which keeps each detail where it happened. After a merge it is
placed right after the merged record and pays its full length. Fit-or-degrade visits the candidates in order and gives
each the highest resolution that still fits. For the $\ell$-th candidate $u_{(\ell)}$,
\begin{equation}
  \begin{aligned}
  v_{(\ell)} &= \max_{\rec \prec \exc \prec \raw} \bigl\{ v : \mathrm{cost}_{(\ell)}(v) \le R_{\ell-1} \bigr\},\\
  R_\ell &= R_{\ell-1} - \mathrm{cost}_{(\ell)}\bigl(v_{(\ell)}\bigr).
  \end{aligned}
  \label{eq:fod}
\end{equation}
The record always fits because it costs nothing. A unit whose raw text does not fit keeps the part of it that matches
the query instead of falling back to its record. A more relevant unit is never degraded to make room for a less
relevant one. The rule has no setting tied to the budget, yet a smaller allowance turns more of the recovered units into excerpts.

An excerpt is the record $s_i$ followed, in their original order, by the pieces of $u_i$ that best match the query.
\method ranks the pieces by their Okapi BM25 (BM25) score against the query
(Table~\ref{tab:queries}), with inverse document frequencies over the whole store, and takes them greedily into a set $X_i$ while $\sum_{\pi \in X_i} |\pi| \le E$. Excerpts are verbatim because a later call often needs exact strings.

\subsection{Piece Recovery}
\label{sec:pieces}

Piece recovery spends the room on pieces from the whole history instead of whole units. It ranks every piece of every
unit, $\Pi = \bigcup_i \Pi_i$, by its BM25 score against the query, which is the question for a reader and the task with
the newest unit for an agent. Pieces enter a set $X_t$ in rank order while $\sum_{\pi \in X_t} |\pi| \le R_0$. The chosen pieces follow the timeline, grouped by unit in time order. Because
pieces always want more room than is left, the timeline is merged down to its share $\rho A_t$ and lends any unused part of it to the pieces. One unit can contribute several pieces and many units one each, which suits evidence spread over sessions.

\subsection{Choosing a Recovery Mode}
\label{sec:choice}

A question about a past conversation usually names its evidence, and keyword matching over all pieces finds it while
spending the room only on matching text. The next step of an agent re-uses a detail that its context cannot name, and
whole units keep the context around each detail. For this reason we use unit recovery for agents and piece recovery
for questions over a history. Within a mode \method adapts the detail it shows to the budget at every call, but it does not
choose the mode at run time (Section~\ref{sec:findability}). The only LLM calls at read time write merged records, and
Algorithm~\ref{alg:render} in Appendix~\ref{app:impl} gives the full procedure of unit recovery.

\section{Experimental Setup}
\label{sec:setup}

\begin{table*}[t]
\centering
\footnotesize
\setlength{\tabcolsep}{2.8pt}
\begin{tabular}{lccccccccccccc}
\toprule
 & \multicolumn{4}{c}{AppWorld} & \multicolumn{2}{c}{SWE-bench} & \multicolumn{4}{c}{LongMemEval} & \multicolumn{3}{c}{LoCoMo} \\
\cmidrule(lr){2-5}\cmidrule(lr){6-7}\cmidrule(lr){8-11}\cmidrule(lr){12-14}
 & \multicolumn{2}{c}{Normal} & \multicolumn{2}{c}{Challenge} & \multicolumn{2}{c}{issues at 8k} & & & & & & & \\
\cmidrule(lr){2-3}\cmidrule(lr){4-5}\cmidrule(lr){6-7}
Method & 4k & 6k & 4k & 6k & 95 & 480 & 4k & 8k & 16k & 32k & 2k & 4k & 8k \\
\midrule
Full history & \multicolumn{2}{c}{78.6} & \multicolumn{2}{c}{62.1} & 69.5 & 67.9 & \multicolumn{4}{c}{88.4} & \multicolumn{3}{c}{83.4} \\
\midrule
Truncation & 26.2 & 55.4 & 8.6 & 22.5 & 63.2 & 62.3 & 10.9 & 15.0 & 24.1 & 35.5 & 14.0 & 25.9 & 39.6 \\
Masking \citep{lindenbauer2025complexity} & 35.7 & 60.1 & 9.6 & 27.3 & \second{67.4} & \second{64.2} & -- & -- & -- & -- & -- & -- & -- \\
Masking + rule log & \third{58.3} & 66.7 & \third{27.6} & 42.7 & -- & -- & -- & -- & -- & -- & -- & -- & -- \\
Piece retrieval & 20.2 & 48.8 & 7.2 & 14.1 & 63.2 & 60.4 & \first{78.4} & \first{81.6} & \first{85.7} & 85.9 & \second{67.7} & \second{72.2} & \second{75.5} \\
Unit retrieval & 28.6 & 48.2 & 8.6 & 17.7 & 64.2 & \third{62.7} & \third{54.8} & \third{73.6} & 83.6 & \third{86.6} & \third{48.6} & \third{63.4} & \third{74.8} \\
A-Mem \citep{xu2025amem} & 25.0 & 48.2 & 9.4 & 16.3 & \third{65.3} & -- & 52.0 & 71.6 & \third{84.5} & \first{88.6} & 45.5 & 55.8 & 68.8 \\
\sumret \citep{wang2025openhands} & 54.2 & \third{68.5} & 26.1 & 43.6 & 64.2 & 62.5 & 50.7 & 70.2 & 82.3 & \second{86.8} & 44.9 & 56.9 & 69.5 \\
MemGAS \citep{xu2026memgas} & 10.7 & 26.2 & 3.1 & 9.1 & 58.9 & -- & 37.7 & 52.7 & 65.9 & 70.2 & 27.1 & 39.5 & 52.5 \\
Sculptor \citep{li2026sculptor} & 26.2 & 47.6 & 8.2 & 21.3 & 62.1 & -- & 33.9 & \second{75.9} & 76.4 & 74.5 & 17.1 & 29.2 & 51.6 \\
ReadAgent \citep{lee2024readagent} & \second{61.9} & \second{72.0} & \second{28.8} & \second{54.7} & \second{67.4} & -- & 23.0 & 35.2 & 53.4 & 58.2 & 27.3 & 49.4 & 72.1 \\
AFM \citep{cruz2025afm} & 55.4 & \second{72.0} & 26.1 & \third{52.3} & -- & -- & 37.5 & 58.9 & 75.5 & 76.6 & 45.0 & 60.0 & 62.7 \\
\midrule
\method, unit recovery & \first{64.3} & \first{78.0} & \first{40.3} & \first{62.4} & \first{71.6} & \first{66.5} & \unused{69.5} & \unused{75.0} & \unused{81.4} & \unused{86.1} & \unused{52.5} & \unused{65.2} & \unused{75.7} \\
\method, piece recovery & \unused{58.3} & \unused{70.8} & \unused{30.2} & \unused{52.8} & \unused{64.2} & -- & \second{78.0} & \first{81.6} & \second{85.0} & 85.7 & \first{69.9} & \first{73.2} & \first{78.1} \\
\method, timeline only & 63.7 & 67.9 & -- & -- & 69.5 & -- & 41.4 & 62.7 & 70.5 & 70.5 & 40.3 & 63.8 & 67.8 \\
\bottomrule
\end{tabular}
\vspace{-5pt}
\caption{Main results, completion on AppWorld, resolve rate on SWE-bench, accuracy on 440 LongMemEval questions and
J on LoCoMo. Blue, orange and green mark the best, second and third method with a budget per column. Gray numbers,
from the mode \method does not use on that task, and timeline only are not ranked. Appendix~\ref{app:details} lists
the results from separate runs.}
\label{tab:main}
\vspace{-5pt}
\end{table*}

\subsection{Benchmarks}
\label{sec:benchmarks}

AppWorld \citep{trivedi2024appworld} asks agents to complete everyday tasks through the APIs of nine apps. We report the share of tasks whose unit tests all pass, with 60 steps and budgets of 4k and 6k tokens. Test-Normal
has 168 tasks, and Test-Challenge has 417 harder ones on which no setting was chosen. Without a
budget the longest prompt on Test-Normal has a median of 10.4k tokens. The 147 unused tasks of the Train and Dev sets
confirm the design, with the method, predictions and tests fixed before any run (Appendix~\ref{app:confirm}).
SWE-bench Verified \citep{jimenez2024swebench,openai2024sweverified} asks an agent in mini-swe-agent
\citep{minisweagent} to resolve issues in Python repositories within 250 steps. We report the resolve rate at 8k
on the 480 issues our images can grade, 95 of which were used to choose settings.

LongMemEval \citep{wu2025longmemeval} gives each of 500 questions a history of about 48 sessions and 103k tokens, which
a reader answers from 4k to 32k tokens. We choose settings on 60 development questions and report accuracy on the
other 440. Their aggregate results were inspected while we designed piece recovery and the rule for choosing a mode.
For this reason we treat LongMemEval as comparative evidence and the unused AppWorld tasks as the test of the frozen
design. LoCoMo \citep{maharana2024locomo} has 1,540 questions on 10 conversations in its first four categories,
answered from 2k to 8k tokens with the reader and judge prompts of Mem0 \citep{chhikara2025mem0}, and \method keeps its LongMemEval settings.

\subsection{Baselines and Implementation}
\label{sec:baselines}

All methods share the units, the embedding model, the budget, the token counter and the LLM, GPT-5.4 mini. A-Mem \citep{xu2025amem} fills the allowance with retrieved notes in raw text, and \sumret adds a running LLM summary \citep{wang2025openhands} that takes half of the allowance on
LongMemEval and 70\% on the agentic tasks. MemGAS \citep{xu2026memgas} filters granularities with an LLM, Sculptor \citep{li2026sculptor} manages the
context through tools, and ReadAgent \citep{lee2024readagent} replaces gists with pages chosen by an LLM. AFM
\citep{cruz2025afm}, which we reproduce and calibrate, gives each unit a full, compressed or placeholder form
(Appendix~\ref{app:afm}). Piece and unit retrieval rank pieces by BM25 or units by embedding and write no memory. Observation masking \citep{lindenbauer2025complexity} hides old tool
outputs, alone or with a rule-written log of past steps, and truncation keeps the newest units
(Appendix~\ref{app:baselines}). We tuned \method on the LongMemEval development questions only and kept default values on the agentic tasks
(Appendix~\ref{app:impl}). Claude Sonnet 4.6 and GPT-5.4 judge the answers, and the exact McNemar test compares paired outcomes.

\section{Results}
\label{sec:results}

\subsection{Agentic Tasks}
\label{sec:agentic}

\begin{table}[t]
\centering
\footnotesize
\setlength{\tabcolsep}{2.5pt}
\begin{tabular}{lccc}
\toprule
Setting & \method & Baseline & Gap \\
\midrule
\multicolumn{4}{l}{AppWorld Test-Normal} \\
8k & 75.0 & A-Mem 67.3 & $+$7.7 \\
4k, 150 steps & 80.4 & masking 33.9 & $+$46.4 \\
4k, second run & 65.5 & masking 38.1 & $+$27.4 \\
6k, second run & 74.4 & masking 59.5 & $+$14.9 \\
\midrule
\multicolumn{4}{l}{SWE-bench Verified, 8k} \\
other 385 issues & 65.2 & masking 63.4 & $+$1.8 \\
no network & 67.4 & w/o excerpts 63.2 & $+$4.2 \\
\bottomrule
\end{tabular}
\vspace{-5pt}
\caption{\method against the strongest baseline run in further settings.}
\label{tab:robust}
\vspace{-5pt}
\end{table}

With unit recovery, \method completes more AppWorld tasks than every baseline that works under a budget
(Table~\ref{tab:main}).
The margin is widest on Test-Challenge, where it outperforms the second-best method, ReadAgent, by 11.5\% at 4k
($p < 0.001$) and 7.7\% at 6k ($p = 0.002$). At 6k it also matches the full history, 62.4\% against 62.1\% ($p = 1$).
Test-Normal separates the methods less clearly. Its leads of 2.4\% and 6.0\% over ReadAgent are not significant and
lie within the 4.8\% that separates two runs of the same arm. On all 480 SWE-bench Verified issues the margins are smaller, as \method outperforms the second-best method,
observation masking, by 2.3\% and \sumret by 4.0\% with one fifth of the tokens of \sumret. Neither a rerun of AppWorld nor a limit of 150 steps
removes the significant lead, and on the 385 SWE-bench issues not used to choose settings the lead is 1.8\% (Table~\ref{tab:robust} and Appendix~\ref{app:full}).

\subsection{Confirmation on Unused Tasks}
\label{sec:confirm}

\begin{table}[t]
\centering
\footnotesize
\setlength{\tabcolsep}{4pt}
\begin{tabular}{lcccc}
\toprule
 & \multicolumn{2}{c}{Completion} & \multicolumn{2}{c}{Steps} \\
\cmidrule(lr){2-3}\cmidrule(lr){4-5}
Method & 4k & 6k & 4k & 6k \\
\midrule
Full history & \multicolumn{2}{c}{77.6} & \multicolumn{2}{c}{16.0} \\
\midrule
Observation masking & 38.8 & 70.7 & 41.5 & 24.1 \\
\sumret & 64.6 & 74.1 & 34.8 & 24.4 \\
Timeline only & 76.2 & 79.6 & 32.2 & 28.7 \\
\method, piece recovery & 70.7 & 81.0 & 28.8 & 20.9 \\
\method, unit recovery & 76.9 & 81.6 & 26.9 & 20.7 \\
\bottomrule
\end{tabular}
\vspace{-5pt}
\caption{Confirmation on 147 unused AppWorld tasks, with steps per episode. Timeline only drops recovery.}
\label{tab:confirm}
\end{table}

The 147 unused tasks confirm the AppWorld results (Table~\ref{tab:confirm}). Before the first episode we froze the
code, six arms, two budgets and five predictions. At 4k \method outperforms the strongest baseline, \sumret, by
12.2\% (Holm-adjusted $p = 0.006$), and it differs from the full history by $-$0.7\% and $+$4.1\%, neither
significant. Three of the five predictions hold. The lead over \sumret at 6k misses after correction ($p = 0.061$),
and the lead over the timeline alone at 6k shrinks to 2.0\% ($p = 0.345$). This last miss is a first sign that the
timeline does most of the work (Section~\ref{sec:timeline-carries}).

\subsection{Fact Recall}
\label{sec:factrecall}

Keyword retrieval over the raw history is the strongest simple method for fact recall, and piece recovery keeps pace
with it (Table~\ref{tab:main}). The two stay within 0.7\% of each other at every LongMemEval budget, and at 4k piece
recovery outperforms A-Mem by 25.9\%. That gap closes as the budget grows, until A-Mem leads at 32k with 88.6\%
against 85.7\%. LoCoMo favors piece recovery slightly, by 1.0\% to 2.6\% over piece retrieval. These results do not show that records written by an LLM help fact recall, because removing the timeline changes
LongMemEval accuracy by at most 3.6\%. What they show is that one store serves both kinds of scenario. The memory whose timeline carries agent completion matches a dedicated retriever on questions that name their
evidence, whereas every baseline memory built on text written by an LLM falls behind once the budget is tight.

\subsection{Cost and Other Models}
\label{sec:cost}

\begin{figure}[t]
  \centering
  \includegraphics[width=\columnwidth]{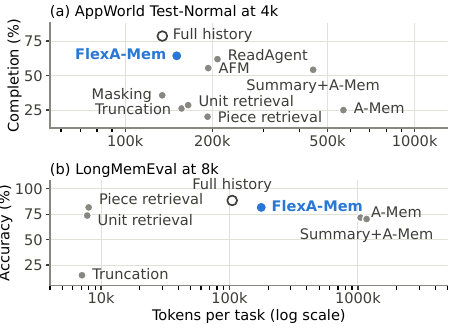}
  \vspace{-22pt}
  \caption{Completion or accuracy against tokens per task, memory construction included. \method uses unit recovery
  on AppWorld and piece recovery on LongMemEval, the mode set for each.}
  \vspace{-5pt}
  \label{fig:cost}
\end{figure}

\begin{table}[t]
\centering
\footnotesize
\setlength{\tabcolsep}{5pt}
\begin{tabular}{lccc}
\toprule
Method & GPT & Gemini & Haiku \\
\midrule
\multicolumn{4}{l}{AppWorld Test-Normal, agent, 4k} \\
\quad Full history & 78.6 & 94.6 & 79.2 \\
\quad Truncation & \third{26.2} & 67.3 & 45.2 \\
\quad Observation masking & \second{35.7} & \second{87.5} & \second{58.3} \\
\quad Piece retrieval & 20.2 & \third{74.4} & \third{53.6} \\
\quad \method w/o excerpts & 64.3 & 91.1 & 67.3 \\
\quad \method & \first{64.3} & \first{94.0} & \first{67.3} \\
\midrule
\multicolumn{4}{l}{LongMemEval, reader, 8k} \\
\quad Truncation & 14.8 & 13.8 & 12.8 \\
\quad A-Mem & \third{73.0} & 75.4 & \second{74.2} \\
\quad \sumret & 71.2 & \third{76.0} & \third{71.4} \\
\quad Piece retrieval & \first{81.2} & \first{86.6} & -- \\
\quad \method w/o excerpts & 74.6 & 77.0 & 74.0 \\
\quad \method & \second{76.0} & \second{81.6} & \first{78.4} \\
\bottomrule
\end{tabular}
\vspace{-8pt}
\caption{AppWorld with three agents and LongMemEval with three readers.}
\label{tab:backbones}
\end{table}

\begin{table*}[t]
\centering
\small
\setlength{\tabcolsep}{5pt}
\begin{tabular}{lccccccc}
\toprule
 & \multicolumn{2}{c}{AppWorld} & SWE-bench & \multicolumn{4}{c}{LongMemEval} \\
\cmidrule(lr){2-3}\cmidrule(lr){5-8}
Variant & 4k & 6k & 8k & 4k & 8k & 16k & 32k \\
\midrule
\method (control) & 65.5 & 76.2 & 66.3 & 69.5 & 75.0 & 81.4 & 86.1 \\
\midrule
without timeline & $-$15.5 & $-$10.1 & $-$1.1 & $+$0.5 & $+$3.6 & $+$1.1 & $-$0.9 \\
without recovery & $-$1.8 & $-$8.3 & $+$3.2 & $-$28.2 & $-$12.3 & $-$10.9 & $-$15.7 \\
without excerpts & $+$0.0 & $+$2.4 & $-$3.2 & $-$15.7 & $-$1.1 & $-$1.4 & $-$1.1 \\
random candidates & $-$0.6 & $-$1.8 & -- & $-$47.5 & $-$40.9 & $-$32.5 & $-$17.0 \\
same selection after the timeline & $+$3.6 & $-$3.6 & -- & $-$0.5 & $+$1.4 & $+$0.5 & $+$0.7 \\
fixed-length recovery & $-$0.6 & $-$4.8 & -- & $+$6.1 & $+$7.0 & $+$1.6 & $-$1.8 \\
\midrule
A-Mem with fit-or-degrade & $+$10.7 & $+$7.7 & -- & $+$13.4 & $+$5.7 & $+$2.0 & $-$1.1 \\
\bottomrule
\end{tabular}
\vspace{-5pt}
\caption{Change in completion or accuracy (\%) when one part of \method with unit recovery is removed or changed. Appendix~\ref{app:details} gives the control of each row.}
\label{tab:ablation}
\vspace{-5pt}
\end{table*}

\method spends one seventh to one third of the tokens of A-Mem and \sumret per task (Figure~\ref{fig:cost} and
Table~\ref{tab:cost}), although the full history costs fewer tokens in every setting. On AppWorld only observation
masking uses fewer tokens among the methods with a budget, and it completes 28.6\% fewer tasks. On LongMemEval piece
retrieval reaches the accuracy of \method with 8.0k instead of 176.6k tokens, because it writes no memory.
On LongMemEval it stays the most accurate method that writes memory with all three readers (Table~\ref{tab:backbones}). Only on SWE-bench Verified with Haiku does \method trail, resolving 2.1\% fewer issues than observation masking
(Appendix~\ref{app:backbones}).

\section{Analysis}
\label{sec:analysis}

\begin{figure*}[t]
  \centering
  \includegraphics[width=0.8\textwidth]{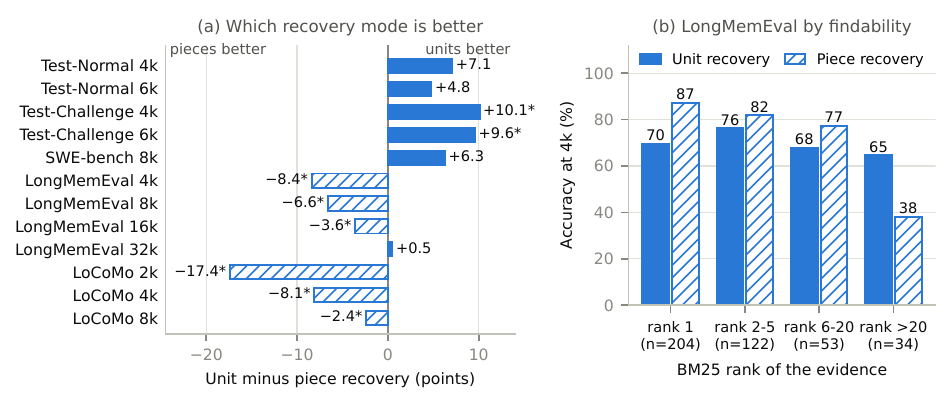}
  \vspace{-15pt}
  \caption{(a) Unit minus piece recovery (\%). (b) LongMemEval accuracy at 4k, grouped by the BM25 rank of the best evidence piece of each question.}
  \vspace{-5pt}
  \label{fig:recovery}
\end{figure*}

\subsection{How the Baselines Fail on AppWorld}
\label{sec:fail}

\begin{figure}[t]
  \centering
  \includegraphics[width=0.8\columnwidth]{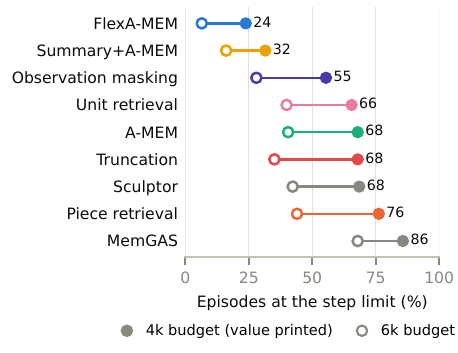}
  \vspace{-15pt}
  \caption{Test-Normal episodes that end at the 60-step limit at 4k (filled, value printed) and 6k (open).}
  \label{fig:steplimit}
\end{figure}

In the failure trajectories we analyze, the main failure mode of the baselines on AppWorld is repeating their own
setup, not giving wrong answers (Figure~\ref{fig:steplimit}). At 4k on Test-Normal, 23.8\% of the episodes of \method
end at the step limit, against 55.4\% to 85.7\% for every baseline except \sumret. The tasks that the baselines declare
complete are right about as often as those of \method. Of their failures at 4k, 90\% are loops of logging in and
rereading documentation. Extra steps do not rescue them. With 150 steps instead of 60, observation masking, truncation and piece retrieval
complete no more tasks, whereas \method rises from 64.3\% to 80.4\%. ReadAgent and AFM, the two strongest baselines, keep a gist of
every unit or a compressed summary of most units in view. Unlike ReadAgent, AFM packs units oldest first and drops 2.3
of the newest units per call at 4k.

\subsection{The Timeline Keeps the Trace}
\label{sec:timeline-carries}

\begin{table}[t]
\centering
\footnotesize
\setlength{\tabcolsep}{4pt}
\begin{tabular}{llcc}
\toprule
Records written by & Recovery & 4k & 6k \\
\midrule
GPT-5.4 mini & none & 67.3 & 75.0 \\
fixed rules, no LLM & none & 61.3 & 61.3 \\
fixed rules, with masking & none & 58.3 & 66.7 \\
Claude Haiku 4.5 & none & 51.8 & -- \\
\midrule
GPT-5.4 mini & raw text only & 64.3 & 80.4 \\
Claude Haiku 4.5 & raw text only & 62.5 & -- \\
\bottomrule
\end{tabular}
\vspace{-5pt}
\caption{AppWorld Test-Normal completion with records from different writers.}
\label{tab:timeline}
\end{table}

Removing the timeline costs the most on AppWorld, lowering completion by 15.5\% at 4k and 10.1\% at 6k
(Table~\ref{tab:ablation}). On SWE-bench Verified the resolve rate barely moves, but the agent needs 56.9 steps per
issue instead of 43.8. The unused tasks tell the same story, with the timeline alone within 2.0\% of the full method
(Table~\ref{tab:confirm}). Records from an LLM help little at 4k. A trace written by fixed rules, with no LLM call,
still completes 61.3\% of Test-Normal against 67.3\% for records from GPT-5.4 mini ($p = 0.15$), although LLM records
matter once there is room at 6k (Table~\ref{tab:timeline}). Such a trace is a strong design on its own. Observation
masking with this trace reaches 58.3\% at 4k, up from 35.7\%, and trails \method by 6.0\% ($p = 0.12$). The gap grows
to 11.3\% at 6k ($p = 0.003$) and on Test-Challenge to 12.7\% at 4k and 19.7\% at 6k (both $p < 10^{-8}$). Fact recall, in contrast, barely uses the timeline. Removing it changes LongMemEval accuracy by at most 3.6\%, and on LoCoMo the timeline alone stays below piece recovery at each budget from 2k to 8k.

\subsection{Recovery Brings Back the Detail}
\label{sec:recovery-adds}

\begin{figure}[t]
  \centering
  \includegraphics[width=0.8\columnwidth]{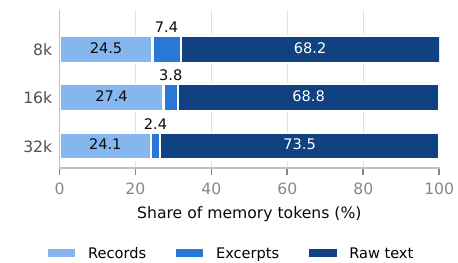}
  \vspace{-10pt}
  \caption{Share of memory tokens of \method on records, excerpts and raw text.}
  \label{fig:composition}
\end{figure}

\begin{table}[t]
\centering
\footnotesize
\setlength{\tabcolsep}{3.5pt}
\begin{tabular}{lccc}
\toprule
Method & Recall & Hit rate & Bal.\ acc. \\
\midrule
Full history & 22/22 & 98.6 & 0.97 \\
Truncation & \second{17/22} & 19.2 & 0.59 \\
Observation masking & \second{17/22} & \first{95.2} & \first{0.96} \\
A-Mem & 11/22 & 25.3 & 0.61 \\
\sumret & \third{12/22} & \third{47.9} & \third{0.73} \\
ReadAgent & \second{17/22} & 32.9 & 0.66 \\
MemGAS & 6/22 & 14.4 & 0.57 \\
\method w/o excerpts & 17/22 & 69.2 & 0.83 \\
\midrule
\method & \first{18/22} & \second{90.4} & \second{0.93} \\
\bottomrule
\end{tabular}
\vspace{-5pt}
\caption{Questions to the memory at 109 checkpoints of SWE-bench trajectories at 8k. Colors mark the best, second
and third method with a budget, without the version without excerpts.}
\label{tab:checkpoint}
\end{table}

Recovery decides accuracy when the answer is an exact detail. On LongMemEval, removing recovery lowers accuracy by
28.2\% at 4k and random candidates lower it by 47.5\%, whereas showing the same selection after the timeline changes
it by at most 1.4\%. Fit-or-degrade is not the best rule there, since giving every candidate an excerpt gains 6.1\%
at 4k. AppWorld uses recovery to save effort rather than to complete more tasks, and on the unused tasks it cuts the steps per
episode from 32.2 to 26.9 at 4k. It still adds 5.0\% on Test-Challenge at 4k through excerpts
($p = 0.031$) and 10.7\% when Claude Haiku 4.5 writes the records (Appendix~\ref{app:timeline}). Without a timeline
the rule decides completion, and rendering the notes of A-Mem with fit-or-degrade raises its Test-Normal completion by
10.7\% at 4k. The rule also adapts to the budget as designed. Excerpts take 7.4\% of the
LongMemEval memory tokens at 8k and 2.4\% at 32k, while raw text grows from 68.2\% to 73.5\%
(Figure~\ref{fig:composition}). With excerpts, the memory tells whether a SWE-bench agent displayed a given file with
a balanced accuracy of 0.93 against 0.83 without them (Table~\ref{tab:checkpoint}). Observation masking reaches 0.96, since it keeps every command and hides only old outputs.

\subsection{When Each Recovery Mode Wins}
\label{sec:findability}

Each mode leads on the kind of scenario it was set for (Figure~\ref{fig:recovery}a). Unit recovery outperforms piece
recovery in all five agentic settings, by 4.8\% to 10.1\%, and piece recovery leads on LongMemEval up to 16k and on
LoCoMo at every budget. In the questions we test, the gap between the modes is associated with how well the question
names its evidence (Figure~\ref{fig:recovery}b). Piece recovery is ahead by 17.6\% at 4k when BM25 ranks the evidence
first, and unit recovery is ahead by 26.5\% when the evidence ranks below twentieth. This rank needs evidence labels.
The runtime signals we tested do not reliably predict the better mode for a single query, and we offer no deployable
rule for choosing it. The two modes also differ in more than the unit of recovery (Appendix~\ref{app:full}).

\section{Conclusion}
\label{sec:conclusion}

A single write-once store, rendered anew at every call as a compact timeline of records plus detail recovered by a
fixed rule, lets \method serve both kinds of scenario. On agentic tasks the timeline
keeps the trace and carries completion, coming within 2.0\% of the full method on tasks unused in design. On fact
recall the recovered text decides accuracy, and keyword retrieval over the kept raw text is already strong. Choosing
the recovery mode for each query without labels is the next step.

\section*{Limitations}

We designed the final \method after the main experiments, and only the 147 unused AppWorld tasks test it with
predictions fixed in advance. Most configurations ran once, with up to 4.8\% between two runs on AppWorld. The
recovery mode is set per task type rather than per query. Every history here fits in the context of the model, where
the full history costs fewer tokens, and all four benchmarks are in English. Appendix~\ref{app:disclosures} gives
further details.

\bibliography{references}
\clearpage
\twocolumn
\tableofcontents

\clearpage
\appendix
\section{Implementation Details}
\label{app:impl}

\begin{algorithm}[h]
\caption{Rendering memory at call $t$.}
\label{alg:render}
\small
\begin{algorithmic}[1]
\REQUIRE notes $\{m_i\}$, timeline $T$, query $q_t$, allowance $A_t$, settings $\rho, \lambda, k, E$
\STATE $\mathcal{C} \gets$ candidates of $q_t$ ordered by $g_t$ (Eq.~\ref{eq:score})
\STATE $D \gets \sum_{j \in \mathcal{C}} \mathrm{cost}_j(\raw)$
\WHILE{$\|T\| > \max(\rho A_t,\ A_t - D)$}
  \STATE merge the oldest records of $T$ into one merged record
\ENDWHILE
\STATE $R \gets A_t - \|T\|$, \quad $\mathcal{V} \gets \emptyset$
\FOR{$j \in \mathcal{C}$ in order}
  \IF{$\mathrm{cost}_j(\raw) \le R$}
    \STATE $\mathcal{V} \gets \mathcal{V} \cup \{(j, \raw)\}$, \quad $R \gets R - \mathrm{cost}_j(\raw)$
  \ELSIF{$\mathrm{cost}_j(\exc) \le R$}
    \STATE $\mathcal{V} \gets \mathcal{V} \cup \{(j, \exc)\}$, \quad $R \gets R - \mathrm{cost}_j(\exc)$
  \ENDIF
\ENDFOR
\STATE $M_t \gets T$ with each $(j, v) \in \mathcal{V}$ in place of the record of $u_j$
\WHILE{$|M_t| > A_t$}
  \STATE remove the lowest ranked element of $\mathcal{V}$ and rebuild $M_t$
\ENDWHILE
\RETURN $M_t$
\end{algorithmic}
\end{algorithm}

\begin{table}[h]
\centering
\footnotesize
\setlength{\tabcolsep}{3pt}
\begin{tabular}{lcc}
\toprule
Setting & LongMemEval & Agentic tasks \\
\midrule
Prompt length $\bar{B}$ & $0.95B$ & $0.95B$ \\
Recent window & none & within $0.25B$ \\
Unit size (tokens) & about 2{,}000 & 1{,}000 \\
Timeline share $\rho$ & 0.3 & 0.5 \\
Discount $\lambda$ & 0.5 & 0.9 \\
Seeds $k$ & 10 & 10 \\
Links per note & 3 & 3 \\
Piece size $\kappa$ & 150 & 150 \\
Excerpt pieces $E$ & 300 & 300 \\
BM25 query & question & newest unit \\
Records merged at once & up to 4 & up to 4 \\
\bottomrule
\end{tabular}
\vspace{-5pt}
\caption{Settings of \method, with tokens counted by the o200k tokenizer.}
\label{tab:hyper}
\end{table}

\begin{table}[h]
\centering
\footnotesize
\setlength{\tabcolsep}{3pt}
\begin{tabular}{lll}
\toprule
Operation & Reader & Agent \\
\midrule
Unit seeds (cosine) & question & task $+$ newest unit \\
Excerpt pieces (BM25) & question & newest unit \\
Piece recovery (BM25) & question & task, newest unit \\
\bottomrule
\end{tabular}
\vspace{-5pt}
\caption{Query of each operation. For seeds, task $+$ newest unit is the normalized sum of their embeddings. BM25
uses the content words of the query.}
\label{tab:queries}
\end{table}

The agent and the reader use medium reasoning effort, and memory calls use low effort. Embeddings come from
all-MiniLM-L6-v2, and tokens are counted with the o200k tokenizer. On LongMemEval the reader receives the benchmark's Chain-of-Note reading prompt \citep{wu2025longmemeval,yu2024chainofnote}
with the memory in place of the history, and the
query is the question. On the agentic tasks the context holds the system prompt and the task, a message with the
memory, and the recent window as the original messages. The recent window is the longest run of newest units within a
quarter of the budget and always contains the newest unit. The query embedding is the normalized sum of the
embeddings of the task and of the newest unit. On agentic tasks the pieces of an excerpt are ranked against the newest
unit alone, because the task text stays the same for the whole episode and would pull every excerpt toward the same
pieces. Table~\ref{tab:queries} lists the query of each operation. Memory starts at the first call that would exceed $\bar{B}$, and below
it every method passes the history through unchanged.

On agentic tasks merged records persist for the whole episode, and merges are sized by the larger of the current
allowance and a reference allowance computed with a full recent window. When the newest unit alone takes more than a
quarter of the budget, the current allowance falls below the reference, and the oldest records are hidden for that
call only. The prompt for a merged record asks for at most 120 words, and longer outputs are cut at 150.

Ties in the BM25 ranking go to the cosine similarity between the piece and the query and then to the earlier piece.
BM25 uses $k_1 = 1.2$ and $b = 0.75$, treats the query as a set of lower-cased content words, and removes stop words and
markup on both sides. Pieces never cross a turn or an output block. A kept piece that does not begin its turn or block
is labeled with its speaker, or with its step number and the first line of its command. Pieces are cut at the first call
after a unit is written, embedded when the unit is first excerpted, and reused at every later call.

Piece recovery shows the timeline first within its share of the allowance, with merges sized as in unit recovery, and
a timeline shorter than its share lends the rest to the pieces. Pieces are ranked as in piece retrieval, by BM25 over
the content words of the query with ties broken by embedding similarity and then by position. They follow the
timeline under a one-line header, grouped by unit in time order with marks for omitted text. The query is the
question on LongMemEval and the task with the newest unit on the agentic tasks. If joining the pieces pushes the
memory past the allowance, it is rendered again with fewer pieces. The control with random candidates makes every
unit a candidate in a random order seeded by a hash of the query, and its excerpts keep the best matches to the query.

\section{Experiment Details}
\label{app:details}

We test paired differences with the exact McNemar test on per-item outcomes. Confidence intervals of differences come
from a paired bootstrap with 10{,}000 resamples, over questions on LongMemEval, over scenarios on AppWorld, over
repositories on SWE-bench and over conversations on LoCoMo. The word significant in the text refers to an uncorrected test at $p < 0.05$, except for the predictions fixed in
advance, which are Holm-corrected. We tested every paired difference in Tables~\ref{tab:robust} and \ref{tab:ablation}
and in Figure~\ref{fig:recovery}a. In Table~\ref{tab:robust} the leads with 150 steps and in the second runs at 4k and
6k are significant. Table~\ref{tab:ablation-full} stars the significant changes of Table~\ref{tab:ablation}, and
Table~\ref{tab:modes} stars the significant gaps of Figure~\ref{fig:recovery}a, which the figure marks in the same way.
Stars in the appendix tables mark $p < 0.05$.
Differences are computed from paired counts before rounding. On LongMemEval the primary judge agrees with the second
on 96.2\% to 99.0\% of the answers in each of the 38 settings of method and budget scored on all 500 questions.

Table~\ref{tab:main} reports completion on the 168 Test-Normal and 417 Test-Challenge tasks of AppWorld, the resolve
rate on the 95 selection issues and all 480 issues of SWE-bench Verified at 8k, accuracy on the 440 held-out
LongMemEval questions under the primary judge, and accuracy (J) on the 1,540 LoCoMo questions of categories 1 to 4.
Column heads give budgets in tokens, full history has no budget, a dash marks a setting that was not run, and masking
is observation masking. Masking does not apply to LongMemEval and LoCoMo, whose conversations hold no tool observations
to mask. Masking + rule log keeps the recent actions of observation masking, turns older outputs into rule-written
outcome lines and keeps a rule-written log of the steps that no longer fit, with no LLM call, and it ran separately on
both AppWorld test sets. Owing to compute cost, the remaining baselines were evaluated on the 95-issue subset of SWE-bench only, and
AFM was not run on SWE-bench. Some AppWorld results in Table~\ref{tab:main} come from runs other than the main run. On both test sets these are
ReadAgent, AFM and masking with a rule log, on Test-Normal also \sumret, MemGAS and Sculptor, and on Test-Challenge
also unit retrieval and piece recovery. Piece recovery on Test-Normal and on the 95 SWE-bench issues comes from a
separate run with its own unit-recovery control, which Table~\ref{tab:modes} compares. AFM uses its better setting on each benchmark, the paper thresholds on
AppWorld and thresholds calibrated on the 60 LongMemEval development questions and on one LoCoMo conversation. Table~\ref{tab:confirm} uses the 147 tasks of AppWorld's Train and Dev sets, timeline only is \method
without recovery, and steps are per episode. Timeline only in Table~\ref{tab:main} is the arm without recovery of
Table~\ref{tab:ablation}, a separate run on AppWorld and SWE-bench, and its LongMemEval values follow from the paired
counts of that table. On LoCoMo it comes from the run of Table~\ref{tab:main}.
On LoCoMo, J is the share of answers that Claude Sonnet 4.6 judges correct with the judge
prompt of Mem0. Table~\ref{tab:cost} counts tokens on both AppWorld test sets at 4k, on the 95
SWE-bench issues (SWE) at 8k and on the 440 held-out LongMemEval questions (LME) at 8k, and \method uses unit recovery. Each LongMemEval question
has its own history, and on LongMemEval piece recovery costs within 0.2k tokens of unit recovery.

Table~\ref{tab:ablation} covers AppWorld Test-Normal, the 95 SWE-bench issues and the 440 held-out LongMemEval
questions. Each row is compared with its own control. On LongMemEval the control of every row is the first row, the run of
Table~\ref{tab:main}, which is also the in-run control of the render controls. On AppWorld and SWE-bench the first row
is a separate run of the full method, and it is the control of the rows without the timeline, without recovery and
without links. The row without excerpts is compared with the run of Table~\ref{tab:main}. Random candidates is compared
with its own control run, which completes 65.5\% and 75.6\%, and the other render controls with a control in their
own run, which completes 63.7\% and 78.6\%. A-Mem with fit-or-degrade is compared with A-Mem, in the same run on
AppWorld (22.0\% and 47.0\%) and with the A-Mem run of Table~\ref{tab:main} on LongMemEval. Random candidates makes every unit a candidate in a random order, the
next row shuffles only the relevant candidates, the next shows the same selection after the timeline, and fixed-length
recovery gives every candidate an excerpt instead of the highest resolution that fits the room.

In Figure~\ref{fig:budget}, ReadAgent and AFM on both AppWorld test sets and \sumret on Test-Normal come from separate
runs, and in Figure~\ref{fig:steplimit} so do
\sumret, Sculptor and MemGAS. Figure~\ref{fig:recovery}a plots the pairs of Table~\ref{tab:modes}, and
Figure~\ref{fig:recovery}b groups the held-out LongMemEval questions by the rank at which BM25 with the bare question
places the best piece of the evidence. In Table~\ref{tab:modes} the last column gives units minus pieces and the
number of items that only unit recovery or only piece recovery solves. Test-Normal, the 95 SWE-bench issues and
LoCoMo pair two arms of one run, and Test-Challenge and LongMemEval pair separate runs. In Table~\ref{tab:types}, SS is single-session, update is knowledge update, and abstention questions also count in their own type.

In Table~\ref{tab:backbones} GPT, Gemini and Haiku are GPT-5.4 mini, Gemini 3.1 Pro and Claude Haiku 4.5, and the
colors leave out the version without excerpts. GPT-5.4 mini
writes the memory in every case, \method uses unit recovery, the answers of Claude Haiku 4.5 are judged by GPT-5.4,
and rows run with only one model are left out. In Table~\ref{tab:models}, Full is the full history, w/o exc.\ is
\method without excerpts, Base is the mean of observation masking, piece retrieval and truncation, and the memory is
the mean of \method and its version without excerpts. GPT-5.4 mini averages two runs. In Table~\ref{tab:robust} the gap is an absolute difference in \%, the rows with 150 steps use 56 tasks, and without network access only \method and its version
without excerpts were rerun on the 95 issues. In Table~\ref{tab:timeline} the first two rows share a run, as do the
two rows with Claude Haiku 4.5, and no row uses excerpts. The row with masking comes from its own run, which keeps the recent actions of observation masking and
turns older outputs into rule-written outcome lines. In Table~\ref{tab:checkpoint}, recall covers 22 questions
about earlier events, and hit rate and balanced accuracy cover 307 questions that ask whether the agent displayed a given file.

In Table~\ref{tab:variants}, raw text first gives raw text to every candidate that fits before any excerpt, and
memory evolution lets an LLM choose the links and rewrite the notes of neighbors in \method without excerpts.
Table~\ref{tab:afm} uses AppWorld Test-Normal, with thresholds of 0.45 and 0.25 in the paper setting and 0.30 and 0.15
after calibration. Figure~\ref{fig:composition} uses all 500 LongMemEval questions. Figure~\ref{fig:case} wraps long lines and removes the user's name and email.

Several further details of the design and the runs bear on the results.\label{app:disclosures} We chose the excerpt length on the 60 development questions of LongMemEval with their evidence labels, and the query
for ranking pieces on agentic tasks on the 95 SWE-bench issues used for selection. In the render control that shows
the selection after the timeline, the added header changes the selection in 10.8\% of the renders. AFM is our
reproduction from its paper, and its calibrated thresholds were chosen on development data. Two runs of the timeline alone differ by 7.1\% at 6k on Test-Normal, and two runs of \method differ by 5.3\% on the
95 SWE-bench issues. The budget is enforced on our own token count, and on the 95 SWE-bench issues the provider billed
more than the budget on 8.9\% of the calls of \method that used memory. Rerunning the five episodes that held most of
these calls with 85\% of the budget removed every such call and left the number of resolved issues unchanged. The
SWE-bench agents could reach the internet, and without network access both \method and its version without excerpts
lose 4\% to 5\%. On LongMemEval each history serves a single question, and the tokens of memory construction
are never shared across questions, and each merged record costs an LLM call at read time.

\section{Baselines}
\label{app:baselines}

A-Mem \citep{xu2025amem} follows the original prompts for note construction, link generation and memory evolution. Each note is indexed
by its metadata and by its raw text in spans of 150 tokens. At read time A-Mem's search order is kept, and raw notes
are taken while they fit and shown in time order. \sumret adds a running summary written with the summarizing prompt
of OpenHands and MemGym \citep{wang2025openhands,xu2026memgym}, extended with an instruction to keep names, numbers, identifiers, paths and error messages. The summary takes
its share of the allowance and lends what it does not use to retrieval. On LongMemEval it is built once over the whole
history before the question is seen. On agentic tasks it is updated as units leave the recent window.

MemGAS \citep{xu2026memgas} keeps every unit as raw text, as turns or steps, as an LLM summary and as LLM keywords. It links units with a
Gaussian mixture over their similarities, weights the granularities per query, ranks units with personalized PageRank
and passes them through one LLM filter. Sculptor \citep{li2026sculptor} shows each unit as a fragment that the model can summarize, fold, restore
or search through the paper's tools, with the tool schemas counted against the budget. ReadAgent \citep{lee2024readagent} writes one gist per
unit and lets one LLM call per question, or per agent call on SWE-bench, choose the units whose raw text replaces
their gists. MemGAS, Sculptor, ReadAgent and
AFM are our implementations from their papers. They run under the same harness, budgets, agent or reader model and
judges as the other methods, and they reuse the prompts written for SWE-bench and LongMemEval on the other agentic and
conversational benchmarks.

Piece retrieval and unit retrieval make no LLM call. Unit retrieval ranks raw units by the best embedding similarity of
their spans of 150 tokens and shows the units that fit in time order. Piece retrieval ranks the pieces of every unit by
BM25 and shows the best pieces that fit, grouped by unit in time order with marks for omitted text. Observation masking
\citep{lindenbauer2025complexity} replaces tool outputs older than the last ten steps with a placeholder once the context would exceed $\bar{B}$, and masks more steps while the context still exceeds it.

\section{Additional Related Work}
\label{app:related}

Several memories fix the unit or the tier of detail when they write. SeCom shows that turn, session and summary
units each fall short and builds segment units denoised by prompt compression \citep{pan2025secom}, yet it commits to
one granularity at write time. MemoryOS moves dialogue from a short-term queue into mid-term segments and a long-term
store, each with its own budget \citep{kang2025memoryos}, and recursive summarization keeps one rolling summary of a
long dialogue \citep{wang2025recursive}. GAM keeps a memo per session and the raw sessions as pages, and an LLM
researcher plans, searches and reflects over them at read time \citep{yan2025gam}, which raises its serving time to
12.43 seconds against 0.52 for A-Mem in that paper. EdgeMem keeps the original turns and builds its index without a
generative LLM call \citep{cui2026edgemem}, but units outside the retrieved set stay invisible. \method keeps the raw
text of every unit, a trace of every unit in the timeline, and a rule that chooses the resolution without an LLM call.

Other work argues for keeping raw text instead of rewriting it. Repeated consolidation by an LLM turns useful memories
faulty, which leads its authors to treat raw episodes as evidence \citep{zhang2026faulty}. Compactors keep on average
17\% of injected constraints, and most do worse than no compaction \citep{wang2026lostincompaction}. Under compression
agents also call retrieval tools far more often, 63.9 times against 21.0 in one study \citep{liu2026compressioncost}.
These findings motivate writing each unit once and recovering its exact text when a call needs it.

Agents can also manage their context themselves. Memex(RL) and GCC let the agent read raw experience back through an
exact key or a context command \citep{wang2026memex,wu2025gcc}, and Pichay restores an evicted tool output only after
the agent issues the same call again \citep{mason2026pichay}. Other agents are trained to edit or condense their own
context \citep{zhang2026memact,liu2026cat}, and AgeMem learns one memory for short and long horizons with
reinforcement learning and tool actions \citep{yu2026agemem}. Recovery then depends on an action of the agent, whereas
\method recovers by a fixed rule before every call.

Coding agents compact their history to fit the budget. SWE-agent replaces old observations with one-line placeholders
\citep{yang2024sweagent}, the OpenHands condensers mask, truncate or summarize \citep{wang2025openhands,wang2026openhandssdk},
and ACON optimizes the guidelines of an LLM compressor \citep{kang2026acon}. Observation masking matches LLM summarization at about half the cost on SWE-bench Verified \citep{lindenbauer2025complexity}, but the removed content cannot return, and repeated compaction erodes even the rules an agent was told to keep \citep{zerhoudi2026compaction}. Other methods reduce what the agent reads. LLMLingua and LLMLingua-2 drop tokens from a single prompt
\citep{jiang2023llmlingua,jiang2024longllmlingua,pan2024llmlingua2}, AttnCompress keeps a fixed share of each
observation as verbatim blocks \citep{zeng2026attncompress}, and AgentDiet removes 39.9\% to 59.7\% of input tokens by
reducing trajectories \citep{xiao2026agentdiet}. Fewer tokens need not mean a lower bill, and one study finds that
cutting tool-output tokens raised the billed cost of the agent \citep{weinberger2026token}. We count every call and
every memory construction in tokens per task. Unlike these methods, \method never removes the raw text of a unit from
its store.

Among other benchmarks, LoCoMo tests very long conversational memory \citep{maharana2024locomo}, and SWE-rebench
collects fresh issues for decontaminated evaluation of coding agents \citep{badertdinov2025swerebench}. MemGym offers a long-horizon memory environment \citep{xu2026memgym}.

\section{Additional Results}
\label{app:results}

\subsection{Confirmation Fixed in Advance}
\label{app:confirm}

The 147 AppWorld Train and Dev tasks had not been used by any run, analysis or design decision, and the agent's
prompt holds no demonstrations. Before the first episode we froze the code, the six arms of Table~\ref{tab:confirm},
the two budgets and five predictions. \method with unit recovery was predicted to lead \sumret at 4k and at 6k, to
lead the timeline alone at 6k, and to be no more than 5\% below piece recovery at 4k and at 6k. The first three
were tested with one-sided exact McNemar tests and the last two with a non-inferiority test on a bootstrap over
scenarios, with Holm correction over all five. The adjusted p-values are 0.006, 0.061, 0.345, 0.0005 and 0.024 in that
order. Power computed beforehand from Test-Normal was 0.36 for the lead over \sumret at 6k and 0.50 for the lead over
the timeline alone. One episode imported AppWorld's internal modules, and it was rerun with the frozen code.

\subsection{Results by Budget}
\label{app:budget}

\begin{figure*}[t]
  \centering
  \includegraphics[width=\textwidth]{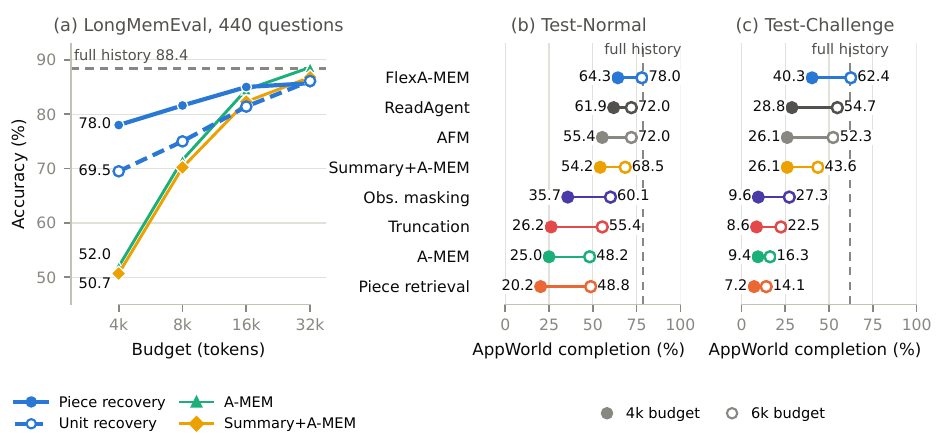}
  \vspace{-5pt}
  \caption{(a) LongMemEval accuracy on 440 held-out questions. (b, c) AppWorld completion rate at 4k
  (filled) and 6k (open). Dashed lines mark the full history.}
  \label{fig:budget}
\end{figure*}

Figure~\ref{fig:budget} plots the results of Table~\ref{tab:main} by budget. On LongMemEval piece recovery leads every
memory built by an LLM at 4k and 8k, and the gaps shrink as the budget grows. On AppWorld the lead of \method over the second-best method is largest at 4k on Test-Challenge, 11.5\%, and smallest
at 4k on Test-Normal, 2.4\%.

\subsection{AFM and LoCoMo}
\label{app:afm}

\begin{table}[t]
\centering
\footnotesize
\setlength{\tabcolsep}{3pt}
\begin{tabular}{lccccc}
\toprule
 & \multicolumn{3}{c}{LoCoMo} & \multicolumn{2}{c}{AppWorld} \\
\cmidrule(lr){2-4}\cmidrule(lr){5-6}
Method & 2k & 4k & 8k & 4k & 6k \\
\midrule
AFM, paper thresholds & 43.8 & 46.0 & 46.2 & 55.4 & 72.0 \\
AFM, calibrated & 45.0 & 60.0 & 62.7 & 48.2 & 57.7 \\
\method, unit recovery & 52.5 & 65.2 & 75.7 & 64.3 & 78.0 \\
\bottomrule
\end{tabular}
\vspace{-5pt}
\caption{AFM with its paper and calibrated thresholds, against \method with unit recovery.}
\label{tab:afm}
\end{table}

We reproduce AFM from its paper. A unit's score multiplies the best embedding similarity of its spans with the query,
a recency term with a half-life and an LLM importance label, and two thresholds map the score to a full, compressed or
placeholder form. Units are packed oldest first, each falls back to the next form when it does not fit, and an LLM
compresses a unit to about 100 tokens once. The prompts and the compression target are our choices. The paper set its
thresholds for another embedder, and with ours the median score is 0.21 while the full-text threshold of 0.45 lies
above the 99th percentile. For this reason we compared four threshold pairs on one LoCoMo conversation with a rule fixed in
advance, the highest mean accuracy over the three budgets, and chose 0.30 and 0.15, which reach 52.6 against 43.0 for
the paper's values.

On LongMemEval the histories score lower, and we compared the paper's thresholds with three pairs on the 60
development questions by mean accuracy over the four budgets, which chose 0.15 and 0.07 (74.2 against 32.5 for the
paper's values). Five threshold pairs were tried in all. On the 440 held-out questions AFM then reaches 37.5, 58.9,
75.5 and 76.6 from 4k to 32k, against 33.2 to 34.1 with the paper's thresholds.

Which AFM setting works better depends on the benchmark (Table~\ref{tab:afm}). Calibration raises AFM on LoCoMo by
about 14\% and 16\% at 4k and 8k, but lowers it on AppWorld. There more units pass the lower thresholds, the full
units fill the room, and oldest-first packing drops 4.4 of the newest units per call at 4k against 2.3 with the paper's
values. \method with unit recovery leads both settings on both benchmarks. Against the better setting it outperforms AFM by 8.9\% and 6.0\% on AppWorld and by 7.5\%, 5.2\% and 13.1\% on
LoCoMo. Without the conversation used for calibration, AFM reaches 45.2\%, 60.3\% and 63.2\% on LoCoMo, and \method
still leads by 7.7\%, 4.5\% and 11.6\%.

On LoCoMo each session forms one unit, and every memory is built once per conversation without its questions. The
timeline alone reaches 40.3, 63.8 and 67.8 at 2k, 4k and 8k, and A-Mem with fit-or-degrade 47.8, 57.1 and 69.1, close
to A-Mem, because a LoCoMo session usually fits whole. Memory construction costs 35.3k tokens per conversation for
\method, against 189.2k for A-Mem and 211.9k for \sumret.

\subsection{Recovery Modes in Detail}

\begin{table}[t]
\centering
\footnotesize
\setlength{\tabcolsep}{2.8pt}
\begin{tabular}{llccc}
\toprule
Benchmark & Budget & Units & Pieces & Units $-$ pieces \\
\midrule
Test-Normal & 4k & 65.5 & 58.3 & $+$7.1 (23/11) \\
 & 6k & 75.6 & 70.8 & $+$4.8 (23/15) \\
Test-Challenge & 4k & 40.3 & 30.2 & $+$10.1$^*$ (67/25) \\
 & 6k & 62.4 & 52.8 & $+$9.6$^*$ (80/40) \\
SWE-bench & 8k & 70.5 & 64.2 & $+$6.3 (9/3) \\
\midrule
LongMemEval & 4k & 69.5 & 78.0 & $-$8.4$^*$ (33/70) \\
 & 8k & 75.0 & 81.6 & $-$6.6$^*$ (25/54) \\
 & 16k & 81.4 & 85.0 & $-$3.6$^*$ (21/37) \\
 & 32k & 86.1 & 85.7 & $+$0.5 (23/21) \\
\midrule
LoCoMo & 2k & 52.5 & 69.9 & $-$17.4$^*$ (133/401) \\
 & 4k & 65.2 & 73.2 & $-$8.1$^*$ (135/259) \\
 & 8k & 75.7 & 78.1 & $-$2.4$^*$ (125/162) \\
\bottomrule
\end{tabular}
\vspace{-5pt}
\caption{Unit and piece recovery paired by task or question. $^*$ marks $p < 0.05$.}
\label{tab:modes}
\end{table}

\begin{table}[t]
\centering
\footnotesize
\setlength{\tabcolsep}{2.5pt}
\begin{tabular}{lcccccc}
\toprule
 & \multicolumn{2}{c}{4k} & \multicolumn{4}{c}{Pieces $-$ units} \\
\cmidrule(lr){2-3}\cmidrule(lr){4-7}
Type & Units & Pieces & 4k & 8k & 16k & 32k \\
\midrule
SS user & 85 & 93 & $+$8.3 & $+$5.0 & $+$5.0 & $+$0.0 \\
SS assistant & 98 & 91 & $-$6.5 & $-$6.5 & $-$2.2 & $+$0.0 \\
SS preference & 85 & 80 & $-$5.0 & $+$5.0 & $-$10.0 & $+$0.0 \\
multi-session & 54 & 62 & $+$7.3 & $+$4.9 & $+$4.9 & $+$0.8 \\
temporal & 60 & 77 & $+$17.1 & $+$9.8 & $+$5.7 & $+$0.8 \\
update & 76 & 85 & $+$8.8 & $+$14.7 & $+$4.4 & $-$5.9 \\
abstention & 48 & 41 & $-$7.4 & $-$11.1 & $-$7.4 & $+$0.0 \\
\midrule
all & 69.5 & 78.0 & $+$8.4 & $+$6.6 & $+$3.6 & $-$0.5 \\
\bottomrule
\end{tabular}
\vspace{-5pt}
\caption{Unit and piece recovery by LongMemEval question type at 4k and their gap at each budget.}
\label{tab:types}
\end{table}

Piece recovery leads on most question types at 4k, by the most on temporal reasoning with 17.1\%
(Table~\ref{tab:types}). Unit recovery leads on questions about what the assistant said, on preference questions and
on the abstention questions. On LongMemEval the two modes cost the same, within 0.2k tokens per question, because piece recovery reuses the notes and merged records that unit recovery writes.

The two modes also reach different evidence. At 4k piece recovery alone shows part of every evidence turn
for 106 questions and wins 54 of the 56 decided there, and unit recovery alone does so for 21 questions and wins 12 of
13.

Piece retrieval and unit recovery fail on different questions. At 8k piece retrieval shows 86.4\% of
the held-out evidence turns of LongMemEval verbatim, against 77.4\% for unit recovery. On the 82 of all 500 questions
where piece retrieval shows only part of the evidence or none of it, unit recovery answers more often. It is correct
on 56.9\% against 41.4\% where piece retrieval shows part of the evidence, and on 62.5\% against 37.5\% where it
shows none.

\subsection{Other Agents and Readers}
\label{app:backbones}

\begin{table}[t]
\centering
\footnotesize
\setlength{\tabcolsep}{3pt}
\begin{tabular}{lccccc}
\toprule
Agent & Full & \method & w/o exc. & Base & Share \\
\midrule
GPT & 77.4 & 64.9 & 65.5 & 27.9 & 75 \\
Gemini & 94.6 & 94.0 & 91.1 & 76.4 & 89 \\
Haiku & 79.2 & 67.3 & 67.3 & 52.4 & 56 \\
\bottomrule
\end{tabular}
\vspace{-5pt}
\caption{Share of the 4k budget loss that the memory recovers on AppWorld Test-Normal, by agent.}
\label{tab:models}
\end{table}

The lead on AppWorld holds with Gemini 3.1 Pro and Claude Haiku 4.5 as agents (Table~\ref{tab:backbones}). With
Gemini, \method completes 94.0\% of the tasks, level with the full history at 94.6\%, and outperforms observation
masking, the second-best method with that agent, by 6.5\%. With Haiku it completes 67.3\% and outperforms observation
masking by 8.9\%. GPT-5.4 mini writes the memory in every case, and only the agent changes.

On LongMemEval, \method is the most accurate method that writes memory at 8k with all three readers and under both
judges. Excerpts matter more with the two other readers, adding 4.6\% with Gemini 3.1 Pro (30 wins, 7 losses) and
4.4\% with Claude Haiku 4.5 (33 wins, 11 losses) but only 1.4\% with GPT-5.4 mini (22 wins, 15 losses, not
significant). At 16k \method ties A-Mem and \sumret with both other readers, with no paired difference above 2.2\%. The answers of Claude Haiku 4.5 are judged by GPT-5.4, because the primary judge comes from the same model family as Haiku.

We measure how much of the loss caused by a 4k budget the memory recovers on AppWorld with each agent
(Table~\ref{tab:models}). The share is the gap between the memory and the mean of three baselines, divided by the gap
between the full history and that mean. The memory recovers 75\% of the loss with GPT-5.4 mini, 89\% with Gemini 3.1
Pro and 56\% with Claude Haiku 4.5. The agents differ in how fast their history fills the budget. The history of
GPT-5.4 mini passes the 4k budget before its first login in 87.5\% of the episodes, against 53.6\% for Gemini 3.1 Pro
and 57.1\% for Claude Haiku 4.5.

On SWE-bench Verified with Claude Haiku 4.5 as the agent, no method with a budget comes close to the agent without
one, which resolves 72.6\% of the 95 issues. \method resolves 61.1\%, against 65.3\% for its version without
excerpts and 63.2\% for observation masking.

\subsection{Token Costs}
\label{app:cost}

\begin{table}[h]
\centering
\footnotesize
\setlength{\tabcolsep}{2.5pt}
\begin{tabular}{lrrrr}
\toprule
 & \multicolumn{2}{c}{AppWorld} & & \\
\cmidrule(lr){2-3}
Method & Normal & Challenge & SWE & LME \\
\midrule
Full history & 134.4 & 218.7 & 210.4 & 104.9 \\
Truncation & 156.7 & 181.9 & 233.5 & 7.1 \\
Masking & 134.3 & 181.6 & 156.2 & -- \\
Piece retrieval & 192.6 & 223.4 & 521.9 & 8.0 \\
Unit retrieval & 165.0 & -- & 459.9 & 7.8 \\
A-Mem & 567.4 & 826.3 & 1{,}148.2 & 1{,}049.5 \\
\sumret & 445.9 & 744.9 & 1{,}540.5 & 1{,}168.3 \\
ReadAgent & 208.4 & 316.5 & 500.0 & -- \\
AFM & 193.6 & 285.6 & -- & -- \\
\midrule
\method & 150.7 & 232.1 & 344.2 & 176.6 \\
\bottomrule
\end{tabular}
\vspace{-5pt}
\caption{Thousands of tokens per task, memory construction included, at 4k on AppWorld and 8k on SWE-bench and
LongMemEval (LME).}
\label{tab:cost}
\end{table}

Table~\ref{tab:cost} lists the costs behind Figure~\ref{fig:cost} and the two settings the figure leaves out,
Test-Challenge at 4k and the 95 SWE-bench issues at 8k. In the figure \method on LongMemEval is plotted at the token
count of unit recovery, which is within 0.2k of that of piece recovery. MemGAS, Sculptor and masking with a rule log
have no cost measurement and are left out.

\subsection{Robustness}
\label{app:robust}

Budgets, step limits, repeated runs and issue sets all leave the lead of \method in place (Table~\ref{tab:robust}).
At 8k on AppWorld it outperforms A-Mem, the strongest baseline run at that budget, by 7.7\%. A second run of AppWorld changes the outcome of 15\% to 17\% of the tasks for \method and keeps every
gap to the rerun baselines significant. On the 385 SWE-bench issues not used to choose settings, \method outperforms observation masking by 1.8\% and \sumret
by 3.1\%. Resampling the LongMemEval reader at 8k changes 19 of the 440 answers of unit recovery but moves its
accuracy by only 0.2\%.

\subsection{What the Timeline Needs}
\label{app:timeline}

A trace written by fixed rules keeps most of the benefit of the timeline at 4k, and records written by an LLM matter
once there is room (Table~\ref{tab:timeline}). The trace is built from the agent's actions without an LLM call
and completes 61.3\% of the Test-Normal tasks at 4k against 67.3\% for records written by GPT-5.4 mini. This difference is not significant, and the trace still outperforms observation masking by 25.6\%. At 6k the trace
stays at 61.3\% while the LLM records reach 75.0\%, which gives a significant gap of 13.7\%.

Recovery also makes the memory tolerant of a weaker writer. With records from Claude Haiku 4.5 the timeline alone
completes 15.5\% fewer tasks than with records from GPT-5.4 mini. Recovery of raw text closes most of that gap. The
memory written by Haiku then completes 62.5\% of the tasks, 1.8\% below the same memory written by GPT-5.4 mini, and
within its run recovery adds 10.7\%.

\subsection{What the Memory Recalls on SWE-bench}
\label{app:checkpoint}

\method keeps track of what a coding agent has seen nearly as well as the full history
(Table~\ref{tab:checkpoint}). We stop the full-history trajectories of the 95 SWE-bench issues at 109 checkpoints and
render each memory at 8k. We then ask the LLM 22 questions about earlier events and 307 questions about whether the
agent displayed a given file. On the second kind \method reaches a balanced accuracy of 0.93, against 0.83 without
excerpts, 0.73 for \sumret and 0.61 for A-Mem. Observation masking reaches 0.96, because it keeps every command and
hides only old outputs.

Re-runs show where the remaining errors come from. When the agent re-runs a command whose output has left its
context, the memory usually held the right unit but lacked the room to show it. For \method without excerpts on the 95 issues, the unit with the earlier output
was a candidate for 84.4\% of 1{,}666 such re-runs. In 39.2\% of them its raw text was larger than the room. Replaying
these steps with excerpts shows part of the earlier output in only 5.8\% of them. In about 70\% of the cases where the
raw text was too long, the room could not hold an excerpt either.

\subsection{How the Memory Uses the Budget}
\label{app:budget-use}

As the budget grows, the memory shifts toward raw text (Figure~\ref{fig:composition}). Excerpts take
7.4\% of the memory's tokens at 8k and 2.4\% at 32k, raw text grows from 68.2\% to 73.5\%, and records stay near a
quarter. At 8k unit recovery shows 77.4\% of the held-out evidence turns verbatim against 71.8\% without excerpts, and
11 of its 20 wins over that version come with more evidence shown, against 2 of its 15 losses.

\subsection{Excerpt Variants and Plug-ins}

\begin{table}[t]
\centering
\small
\setlength{\tabcolsep}{4pt}
\begin{tabular}{lcccc}
\toprule
Variant & 4k & 8k & 16k & 32k \\
\midrule
\method ($E = 300$) & 69.5 & 75.0 & 81.4 & 86.1 \\
excerpts of 150 tokens & 73.2 & 78.2 & -- & -- \\
excerpts of 600 tokens & 64.5 & 74.5 & -- & -- \\
raw text first & 55.5 & 74.8 & -- & -- \\
first pieces instead of best & 70.0 & 76.1 & -- & -- \\
\midrule
A-Mem & 52.0 & 71.6 & 84.5 & 88.6 \\
A-Mem with fit-or-degrade & 65.5 & 77.3 & 86.6 & 87.5 \\
\midrule
with memory evolution & 51.1 & 71.1 & 79.8 & 84.5 \\
\quad and fit-or-degrade & 65.0 & 74.8 & 82.5 & 86.6 \\
\bottomrule
\end{tabular}
\vspace{-5pt}
\caption{Excerpt variants and the fit-or-degrade plug-in on 440 held-out LongMemEval questions.}
\label{tab:variants}
\end{table}

Shorter excerpts help at 4k and 8k (Table~\ref{tab:variants}), whereas giving raw text to every candidate first loses
most of the gain at 4k. Fit-or-degrade raises the accuracy of A-Mem and of the variant with memory evolution by
13.4\% and 13.9\% at 4k and by 5.7\% and 3.7\% at 8k.

\subsection{A Case Study}

\begin{figure*}[t]
\centering
\begin{minipage}{0.98\textwidth}
\footnotesize
\begin{verbatim}
[u1 · step 3] Inspected API docs for phone.show_contact_relationships and
  Venmo endpoints search_friends, add_friend, remove_friend, and search_users.
  (…)
[u2 · step 4] Queried API docs for three apps. Supervisor APIs include
  show_active_task, complete_task, (…)
[u3 · step 5 · excerpt] Looked up API docs for phone.show_contact_relationships
  and venmo.search_friends/add_friend/remove_friend. (…)
[step 5] output of $ print(apis.api_docs.show_api_doc(app_name='phone',
  api_name='show_contact_relationships')): [...] (…)
[u4 · steps 6-8] Queried available supervisor, phone, and Venmo APIs. (…)
[u5 · step 9] Looked up API docs for supervisor.show_account_passwords,
  phone.show_contact_relationships, and Venmo friend/user endpoints. (…)
[u6 · steps 10-11] Retrieved supervisor passwords for amazon, file_system,
  gmail, phone, (…) and venmo. (…)
[u7 · steps 12-14 · excerpt] Loaded API docs for phone.login and venmo.login,
  then successfully logged in (…), both returning Bearer access_token
  dicts. (…)
[step 12] assistant: [...]
print(phone_token)
print(venmo_token)
[... 1 block omitted]
\end{verbatim}
\end{minipage}
\vspace{-5pt}
\caption{Memory of \method at step 16 of an AppWorld task at 4k, verbatim and trimmed. The excerpt of steps 12 to
14 keeps the code that names the access tokens.}
\label{fig:case}
\end{figure*}

Figure~\ref{fig:case} shows how unit recovery keeps an AppWorld agent from repeating its setup. The task is ``Reset
friends on venmo to be the same as my friends in my phone''. In the figure, text after a unit label is the record of
that unit, text after a step label is a verbatim piece of an excerpt, and long lines are wrapped. At step 16 with a 4k
budget, the newest unit takes 1{,}571 tokens and leaves 1{,}712 tokens for memory, which holds five
records and two excerpts. The record of steps 12 to 14 says that both logins returned access tokens but not under
which names, and its excerpt keeps the code that printed them. The agent checks that both tokens still exist, never
logs in again, and submits at step 22. On the same task truncation, A-Mem and piece retrieval stop at the limit of 60
steps after reading API documentation 74 to 109 times.

\subsection{Numbers Reported by Other Papers}

\begin{table*}[t]
\centering
\footnotesize
\setlength{\tabcolsep}{3.5pt}
\begin{tabular}{lp{2.2cm}p{3.4cm}lp{2.9cm}}
\toprule
Method & Backbone & Setting & Reported & Reference in the same paper \\
\midrule
\multicolumn{5}{l}{LongMemEval} \\
MemGAS \citep{xu2026memgas} & GPT-4o mini & reader input cut at 8k and 16k, GPT-4o judge & 59.8, 60.3 & full history 50.6 \\
Nemori \citep{ma2026nemori} & GPT-4.1 mini, GPT-4o mini & context of 3.7k to 4.8k tokens & 74.6, 64.2 & full context 65.6, 55.0 \\
LightMem \citep{fang2026lightmem} & GPT-4o mini & no token budget, GPT-4o mini judge & 64.3 to 68.6 & A-Mem 62.6, full text 56.8 \\
\midrule
\multicolumn{5}{l}{SWE-bench Verified} \\
AttnCompress \citep{zeng2026attncompress} & Gemini 3 Flash & 200 issues, no token budget & 71.5 & full history 72.5, masking 67.5 \\
masking \citep{lindenbauer2025complexity} & Gemini 2.5 Flash & 500 issues, 250 turns, no token budget & 35.6 & raw agent 32.8, summary 36.0 \\
\midrule
\multicolumn{5}{l}{AppWorld Test-Normal} \\
ACON \citep{kang2026acon} & GPT-4.1 mini & compression from 4{,}096 history tokens & 42.3 & no compression 35.7 \\
ACON \citep{kang2026acon} & GPT-5 chat & as above & 62.5, 65.5 & no compression 66.7 \\
\bottomrule
\end{tabular}
\vspace{-5pt}
\caption{Numbers that other papers report in their own settings, not comparable to ours.}
\label{tab:published}
\end{table*}

Table~\ref{tab:published} lists numbers that other papers report on our three benchmarks. Their backbones, judges,
splits, harnesses and budgets differ from ours. Without a budget our agents and reader reach 88.4\% on LongMemEval,
67.9\% on the 480 SWE-bench issues and 78.6\% and 62.1\% on the two AppWorld test sets, the references of Table~\ref{tab:main}.

\section{Supplementary Results}
\label{app:full}

This section gives supplementary results for Sections~\ref{sec:results} and \ref{sec:analysis}, including the complete ablations and the longer form of each analysis.

\begin{table*}[h]
\centering
\small
\setlength{\tabcolsep}{5pt}
\begin{tabular}{lccccccc}
\toprule
 & \multicolumn{2}{c}{AppWorld} & SWE-bench & \multicolumn{4}{c}{LongMemEval} \\
\cmidrule(lr){2-3}\cmidrule(lr){5-8}
Variant & 4k & 6k & 8k & 4k & 8k & 16k & 32k \\
\midrule
\method (control) & 65.5 & 76.2 & 66.3 & 69.5 & 75.0 & 81.4 & 86.1 \\
\midrule
without timeline & $-$15.5$^*$ & $-$10.1$^*$ & $-$1.1 & $+$0.5 & $+$3.6$^*$ & $+$1.1 & $-$0.9 \\
without recovery & $-$1.8 & $-$8.3$^*$ & $+$3.2 & $-$28.2$^*$ & $-$12.3$^*$ & $-$10.9$^*$ & $-$15.7$^*$ \\
without links & $+$1.2 & $-$0.6 & $+$5.3 & $+$1.4 & $+$0.5 & $-$0.7 & $-$3.4$^*$ \\
without excerpts & $+$0.0 & $+$2.4 & $-$3.2 & $-$15.7$^*$ & $-$1.1 & $-$1.4 & $-$1.1 \\
random candidates & $-$0.6 & $-$1.8 & -- & $-$47.5$^*$ & $-$40.9$^*$ & $-$32.5$^*$ & $-$17.0$^*$ \\
same candidates, random order & $+$0.6 & $-$4.8 & -- & $-$46.8$^*$ & $-$37.5$^*$ & $-$26.1$^*$ & $-$14.5$^*$ \\
same selection after the timeline & $+$3.6 & $-$3.6 & -- & $-$0.5 & $+$1.4 & $+$0.5 & $+$0.7 \\
fixed-length recovery & $-$0.6 & $-$4.8 & -- & $+$6.1$^*$ & $+$7.0$^*$ & $+$1.6 & $-$1.8 \\
\midrule
A-Mem with fit-or-degrade & $+$10.7$^*$ & $+$7.7$^*$ & -- & $+$13.4$^*$ & $+$5.7$^*$ & $+$2.0 & $-$1.1 \\
\bottomrule
\end{tabular}
\vspace{-5pt}
\caption{All variants of Table~\ref{tab:ablation}, change in completion or accuracy (\%), where $^*$ marks $p < 0.05$.
Appendix~\ref{app:details} gives the control of each row.}
\label{tab:ablation-full}
\end{table*}

\subsection{Results}

\subsubsection{Agentic Tasks}

\method with unit recovery completes more AppWorld tasks than every baseline with a budget, and its lead is largest on
the harder Test-Challenge (Table~\ref{tab:main}). There it completes 40.3\% of the tasks at 4k and outperforms the
second-best method, ReadAgent, by 11.5\% ($p < 0.001$), and by 7.7\% at 6k ($p = 0.002$). At 6k it completes 62.4\%
against 62.1\% for the agent with its full history, with no significant difference ($p = 1$). Test-Normal is less
decisive. The leads over ReadAgent, 2.4\% at 4k and 6.0\% at 6k, are not significant ($p = 0.64$ and $0.12$), while
the 8.9\% lead over AFM at 4k is ($p = 0.017$). At 6k on Test-Normal \method comes within 0.6\% of the full history. A
second run of \method completes 65.5\% and 74.4\% of Test-Normal (Appendix~\ref{app:robust}), and two runs of the same arm differ by up to 4.8\%, which is more than the 2.4\% lead.

SWE-bench Verified repeats this ordering with smaller margins. Among the methods run on all 480 issues, \method
resolves 66.5\% and outperforms the second-best method, observation masking, by 2.3\% and \sumret by 4.0\%, while
\sumret spends 1{,}577.1k tokens per issue against 306.2k. On the 385 issues not used to choose settings these leads
are 1.8\% and 3.1\%. A-Mem, MemGAS,
Sculptor and ReadAgent ran only on the 95 selection issues, where \method resolves 71.6\% against 67.4\% for ReadAgent
and observation masking and 69.5\% for the full history.

\subsubsection{Confirmation on Unused Tasks}

The AppWorld results hold on the 147 unused tasks, which played no part in the design (Table~\ref{tab:confirm}).
Before the first episode we froze the code, six arms, two budgets and five predictions with their tests. \method
completes 76.9\% of these tasks at 4k and outperforms the strongest baseline, \sumret, by 12.2\% (Holm-adjusted
$p = 0.006$). Its differences from the full history, $-$0.7\% at 4k and $+$4.1\% at 6k, are not significant. Three of the five predictions hold after
Holm correction, the lead over \sumret at 4k and non-inferiority to piece recovery at both budgets ($p = 0.0005$ and
$0.024$). Two predictions fail. The 7.5\% lead over \sumret at 6k gives $p = 0.061$, and the lead over the timeline alone at 6k is 2.0\%
($p = 0.345$) against 8.3\% on Test-Normal. This second miss is the first sign that the timeline does most of the work
on AppWorld, which Section~\ref{sec:timeline-carries} examines.

\subsubsection{Fact Recall}

Keyword retrieval over the raw history is the strongest simple method for fact recall, and piece recovery keeps pace
with it (Table~\ref{tab:main}). Piece retrieval, which writes no memory, and piece recovery stay within 0.7\% of each
other at every LongMemEval budget, because questions that name their evidence are served well by keyword matching.
Both lead every memory built by an LLM at tight budgets. At 4k piece recovery answers 78.0\% of the held-out questions
and outperforms A-Mem by 25.9\% and \sumret by 27.3\%. These gaps shrink as the budget grows, and at 32k A-Mem is the
more accurate, 88.6\% against 85.7\%. On LoCoMo piece recovery leads piece retrieval by 1.0\% to 2.6\% at every budget.
Unit recovery there outperforms every baseline memory built by an LLM, by 6.2\% to 9.4\% over A-Mem and \sumret and by
5.2\% to 13.1\% over AFM with calibrated thresholds. For questions that name their
evidence, the value of \method lies in keeping the raw text verbatim more than in the records that an LLM writes for each
unit.

\subsubsection{Cost and Other Models}



\method spends about one seventh to one third of the tokens of A-Mem and \sumret per task, whose memory construction
includes LLM link generation and memory evolution (Table~\ref{tab:cost}). On Test-Normal it uses about as many tokens
as truncation and observation masking while completing far more tasks. The full history costs fewer tokens than
\method in every column, and the savings of \method are relative to other memories, not to reading the whole history.

Changing the agent leaves the AppWorld lead in place (Appendix~\ref{app:backbones}). With Gemini 3.1 Pro as the agent,
\method completes 94.0\% of Test-Normal at 4k against 94.6\% for the full history and 87.5\% for observation masking,
and with Claude Haiku 4.5 67.3\% against 58.3\%. The lead does not carry over to SWE-bench Verified with Haiku as the
agent,
where \method resolves 61.1\% of the 95 issues against 63.2\% for observation masking and 65.3\% for its own version without excerpts.

\subsection{Analysis}

\subsubsection{How the Baselines Fail on AppWorld}

In the failure trajectories we analyze, the main failure mode of the baselines on AppWorld is repeating their own
setup, not giving wrong answers
(Figure~\ref{fig:steplimit}). At 4k on Test-Normal, 23.8\% of the episodes of \method end at the step limit, against
55.4\% to 85.7\% for every baseline except \sumret. When the five baselines of the main run declare a task complete,
they are right about as often as \method, 77.8\% to 85.0\% against 84.4\%. Of their failures at 4k, 90\% are loops in
which the agent logs in, lists the APIs and reads their documentation again. Extra steps do not rescue them. With 150 steps instead of 60, observation masking, truncation and piece retrieval
complete exactly as many tasks as before, whereas \method rises from 64.3\% to 80.4\%. On Test-Challenge at 4k, 47\% of the episodes of \method reach the limit,
against 57\% for piece recovery, 61\% for \sumret and 84\% to 95\% for the other baselines. 

ReadAgent and AFM, the two strongest baselines on Test-Normal, are also the two that keep something of every unit in
view. ReadAgent keeps a gist of every unit and spends 91.0k memory tokens per episode at 4k against 36.8k for \method. AFM keeps most units
as short compressed summaries, but it packs units oldest first and drops 2.3 of the newest units per call at 4k. The ablations below test whether the trace of the whole history is what decides completion.

\subsubsection{The Timeline Keeps the Trace}

Removing the timeline costs the most on AppWorld (Table~\ref{tab:ablation}). Completion falls by 15.5\% at 4k and
10.1\% at 6k, and on SWE-bench Verified the agent needs 56.9 steps per issue instead of 43.8 while its resolve rate
moves by only 1.1\%. On the unused tasks the timeline alone completes 76.2\% and 79.6\%, within 2.0\% of the full
method (Table~\ref{tab:confirm}). Even a trace written by fixed rules from the agent's actions, with no LLM call,
completes 61.3\% of Test-Normal at 4k against 67.3\% for records written by GPT-5.4 mini ($p = 0.15$). At 6k it falls 13.7\% behind (Appendix~\ref{app:timeline}). 

The trace matters where an agent must know what it has already done, and it does little for questions that name their
answer. Without the timeline, LongMemEval accuracy changes by at most 3.6\%, and its one significant effect is a gain of 3.6\%
at 8k. On LoCoMo the timeline alone answers 40.3\% to 67.8\% of the questions from 2k to 8k, below
piece recovery at every budget (Table~\ref{tab:main}).

\subsubsection{Recovery Brings Back the Detail}

Recovery decides accuracy when the answer is an exact detail. On LongMemEval, removing recovery lowers accuracy by
28.2\% at 4k and removing excerpts lowers it by 15.7\%. Four render controls, each changing one part of unit recovery on the same notes and
budget, show that the choice and order of the recovered units decide accuracy there. Making every unit a candidate in a random order loses 47.5\% at 4k and shuffling the relevant candidates loses 46.8\%,
whereas showing the same selection after the timeline changes accuracy by at most 1.4\%. Fit-or-degrade is not the
best rule for this task, since giving every candidate an excerpt instead gains 6.1\% and 7.0\% at 4k and 8k. Excerpts also raise from 0.83 to 0.93 the balanced accuracy with which the memory tells whether a SWE-bench agent
displayed a given file (Appendix~\ref{app:checkpoint}).

On AppWorld recovery changes effort more than completion, and no render control changes the completion rate of
Test-Normal significantly. Unit recovery takes 26.9 steps per episode on the unused tasks against 32.2 for the timeline alone at 4k, and 20.7
against 28.7 at 6k. Fixed-length recovery takes 3.6 and 5.0 more steps than fit-or-degrade.
Excerpts add 5.0\% on Test-Challenge at 4k ($p = 0.031$). Records from Claude Haiku 4.5 leave the timeline alone 15.5\%
behind, and recovery of raw text then adds 10.7\% on Test-Normal (Appendix~\ref{app:timeline}).

Without a timeline the recovery rule decides the completion rate. Rendering the notes that A-Mem writes with fit-or-degrade raises Test-Normal completion by 10.7\% at 4k and 7.7\% at
6k, and LongMemEval accuracy by 13.4\% at 4k. A-Mem shows 0.87 whole notes per call at 4k and leaves much of its allowance empty, because a retrieved step is
often longer than the room, whereas fit-or-degrade fills that room with about two excerpts. A-Mem with fit-or-degrade still completes 21\% to 33\% fewer tasks than \method.

\subsubsection{When Each Recovery Mode Wins}

Each recovery mode leads on the kind of scenario it was set for (Figure~\ref{fig:recovery}a). Unit recovery
outperforms piece recovery in all five agentic settings, by 4.8\% to 10.1\%, significantly at both budgets of
Test-Challenge. Piece recovery leads on LongMemEval by 8.4\%, 6.6\% and 3.6\% from 4k to 16k, and the two tie at 32k.
LoCoMo shows leads of 17.4\%, 8.1\% and 2.4\% from 2k to 8k. The two
modes also differ in where their text appears and in how much room the timeline can borrow (Limitations), and these
gaps compare two configurations rather than isolate the unit of recovery.

In the questions we test, the gap between the modes is associated with how well the question names its evidence (Figure~\ref{fig:recovery}b). We
measure this by the rank at which BM25 with the bare question places the best evidence piece among the roughly
1{,}040 pieces of a history. For the 204 of the 413 held-out questions with an answer in the history whose evidence ranks first, piece recovery is
ahead of unit recovery by 17.6\% at 4k. Unit recovery is ahead by 26.5\% for the 34 questions whose evidence ranks below twentieth.

This rank needs the evidence labels, and the runtime signals we tested do not reliably predict the better mode for a
single query.
On AppWorld the detail an agent uses next lands among the top five keyword matches with its task and latest steps for
22.4\% of its needs, against 22.2\% under a random order. None of nine task properties correlates with the gap. The BM25 score of the best LongMemEval piece has a Spearman correlation of at most 0.05 in
magnitude with the gap at 4k and 8k. For these reasons we set the mode per task type.

\end{document}